\documentclass[twoside]{article}

\usepackage{hyperref}
\usepackage{url}

\usepackage[utf8]{inputenc} 
\usepackage[T1]{fontenc}    
\usepackage{hyperref}       
\usepackage{url}            
\usepackage{booktabs}       
\usepackage{amsfonts}       
\usepackage{nicefrac}       
\usepackage{microtype}      
\usepackage{xcolor}         

\usepackage{amsmath}
\usepackage{amssymb}
\usepackage{amsthm}
\usepackage{mathtools}

\usepackage{todonotes}

\usepackage[round]{natbib}
\renewcommand{\bibname}{References}
\renewcommand{\bibsection}{\subsubsection*{\bibname}}

\usepackage{autonum}

\usepackage{bibunits}
\usepackage{enumitem}
\usepackage{algorithm}
\usepackage{algorithmic}

\usepackage[preprint]{aistats2025}
\begin{document}

%
\runningtitle{Recurrent Sparse Spectrum Signature Gaussian Processes}

%
\runningauthor{Csaba Tóth, Masaki Adachi, Michael A. Osborne, Harald Obehauser}
\newcommand{\eps}{\epsilon}
\renewcommand{\d}{\,\mathrm{d}}
\renewcommand{\k}{\mathrm{k}}
\newcommand{\p}{\prime}

\newcommand{\abs}[1]{\left\vert #1 \right\vert}
\newcommand{\inner}[1]{\langle #1 \rangle}
\newcommand{\norm}[1]{\left\Vert #1 \right\Vert}
\newcommand{\pars}[1]{\left( #1 \right)}
\newcommand{\bracks}[1]{\left[ #1 \right]}

\newcommand{\curls}[1]{\left\{ #1 \right\}}

\newcommand{\diag}{\operatorname{diag}}

\newcommand{\rff}{\tilde\varphi}
\newcommand{\rfsf}{\tilde\Phi}

\newcommand{\rfdsf}{\tilde\Phi^{\b\lambda}}

\renewcommand{\b}{\mathbf}
\newcommand{\bx}{\mathbf{x}}
\newcommand{\by}{\mathbf{y}}
\newcommand{\bz}{\mathbf{z}}
\newcommand{\bM}{\mathbf{M}}
\newcommand{\bi}{\mathbf{i}}
\newcommand{\bj}{\mathbf{j}}
\newcommand{\bb}{\mathbf{b}}
\newcommand{\bw}{\mathbf{w}}

\newcommand{\bbR}{\mathbb{R}}
\newcommand{\bbQ}{\mathbb{Q}}
\newcommand{\bbN}{\mathbb{N}}
\newcommand{\bbZ}{\mathbb{Z}}
\newcommand{\bbE}{\mathbb{E}}
\newcommand{\bbV}{\mathbb{V}}
\newcommand{\bbP}{\mathbb{P}}

\renewcommand{\c}{\mathcal}
\newcommand{\cD}{\mathcal{D}}
\newcommand{\cE}{\mathcal{E}}
\newcommand{\cF}{\mathcal{F}}
\newcommand{\cG}{\mathcal{G}}
\newcommand{\cH}{\mathcal{H}}
\newcommand{\cL}{\mathcal{L}}
\newcommand{\cM}{\mathcal{M}}
\newcommand{\cN}{\mathcal{N}}
\newcommand{\cP}{\mathcal{P}}
\newcommand{\cR}{\mathcal{R}}
\newcommand{\cU}{\mathcal{U}}

\newcommand{\iid}{\text{i.i.d.}}
\newcommand{\simiid}{\stackrel{\iid}{\sim}}
\newcommand{\bs}{\boldsymbol}

\renewcommand{\given}{\,\vert\,}
\newcommand{\setgiven}{\,:\,}

\newcommand{\KL}[2]{\mathrm{KL}\bracks{#1 \,\middle\Vert\, #2}}
\newcommand{\iss}[2]{\sum_{1 \leq i_1 < \cdots < i_{#1} < #2}}
\newcommand{\issd}[3]{\sum_{\substack{1 \leq i_1 < \cdots < i_{#1} < #2\\1 \leq j_1 < \cdots < j_{#1} < #3}}}

\renewcommand{\mod}{\operatorname{mod}}
\newcommand{\DFT}{\operatorname{DFT}}
\newcommand{\IDFT}{\operatorname{IDFT}}

\newcommand{\kernel}{\mathrm{k}}
\newcommand{\paths}{\mathrm{Paths}}
\newcommand{\seq}{\mathrm{Seq}}
\newcommand{\onevar}{{1\text{-var}}}

\newcommand{\spc}{\hspace{3pt}}

\twocolumn[

\aistatstitle{Learning to Forget: Bayesian Time Series Forecasting using\\ Recurrent Sparse Spectrum Signature Gaussian Processes}

\aistatsauthor{ Csaba Tóth$^1$ \And Masaki Adachi$^{2,3}$ \And Michael A. Osborne$^2$ \And Harald Oberhauser$^1$}

\aistatsaddress{
$^1$Mathematical Institute, University of Oxford\\
$^2$Machine Learning Research Group, University of Oxford\\
$^3$Toyota Motor Corporation\\
} ]

\begin{abstract}
  The signature kernel is a kernel between time series of arbitrary length and comes with strong theoretical guarantees from stochastic analysis. It has found applications in machine learning such as covariance functions for Gaussian processes.
  A strength of the underlying signature features is that they provide a structured global description of a time series. However, this property can quickly become a curse when local information is essential and forgetting is required; so far this has only been addressed with ad-hoc methods such as slicing the time series into subsegments.
  To overcome this, we propose a principled, data-driven approach by introducing a novel forgetting mechanism for signatures.
  This allows the model to dynamically adapt its context length to focus on more recent information. 
  To achieve this, we revisit the recently introduced Random Fourier Signature Features, and develop Random Fourier Decayed Signature Features (RFDSF) with Gaussian processes (GPs).
  This results in a Bayesian time series forecasting algorithm with variational inference, that offers a scalable probabilistic algorithm that processes and transforms a time series into a joint predictive distribution over time steps in one pass using recurrence. For example, processing a sequence of length $10^4$ steps in $\approx 10^{-2}$ seconds and in $< 1\text{GB}$ of GPU memory. We demonstrate that it outperforms other GP-based alternatives and competes with state-of-the-art probabilistic time series forecasting algorithms.
\end{abstract}

\section{Introduction}
Time series forecasting plays a central role in various domains, including finance \citep{sezer2020financial}, renewable energy \citep{wang2019review, adachi2023bayesian}, and healthcare \citep{bui2018time}. Despite its significance, ongoing research faces several challenges, such as nonlinear data domains, ordered structure, time-warping invariance, discrete and irregular sampling, and scalability.
Among the numerous existing approaches, we focus on the signature approach, which address the first four challenges and have demonstrated state-of-the-art performance as a kernel for sequential data \citep{toth2023random}. However, the trade-off between accuracy and scalability remains unresolved. While the exact computation of the signature kernel requires quadratic complexity for sequence length $T$ \citep{kiraly2019kernels}, existing approximate methods reduce complexity to linear, at the cost of degraded performance on large-scale datasets. Moreover, these methods treat the signature as a global feature, which limits their ability to capture recent local information effectively. This is particularly crucial for time series forecasting, where capturing near-term correlations are often as important as identifying long-term trends.

\textbf{Contributions.} 
We propose Random Fourier Decayed Signature Features with Gaussian Processes, which dynamically adjust its context length based on the data, prioritizing more recent information, and call the resulting model Recurrent Sparse Spectrum Signature Gaussian Process (RS\textsuperscript{3}GP). By incorporating variational inference, our model learns the decay parameters in a data-driven manner, allowing it to transform time-series data into a joint predictive distribution at scale while maintaining accuracy. Our approach outperforms other GP baselines for time series on both small and large datasets and demonstrates comparable performance to state-of-the-art deep learning diffusion models while offering significantly faster training times.


\vspace{-0.5em}
\subsection{Related work}
\vspace{-0.5em}
\textbf{Signature approach.} 
To address the quadratic complexity of the signature kernel \citep{kiraly2019kernels}, two main approaches have been proposed: (1) subsampling: \citet{kiraly2019kernels} introduced sequence-subsampling via Nyström approximation \citep{williams2000using}, \citet{toth2021seqtens} proposed inter-domain inducing points, and \citet{lemercier2021siggpde} diagonalized the Gram matrix. However, these methods can result in crude approximations and performance degradation on large-scale datasets. (2) random projection, including nonlinear projection techniques \citep{lyons2017sketching, morrill2020generalised}, and Random Fourier Features \citep{rahimi2007random}; \citet{toth2023random} introduced Random Fourier Fignature Features for scalable computation with theoretical guarantees for approximating the signature kernel with high probability. However, all these methods treat the signature as a global feature map and cannot capture recent local or recent information effectively.

\textbf{Gaussian process approach.} 
There are two prominent approaches: (1) specialized kernels for sequences \citep{lodhi2002text, cuturi2011fast, cuturi2011autoregressive, al2017learning}, and (2) state-space models \citep{frigola2013bayesian, frigola2014variational, mattos2016recurrent, eleftheriadis2017identification, doerr2018probabilistic, ialongo2019overcoming}. These approaches can complement each other; modeling the latent system as a higher-order Markov process allows sequence kernels to capture the influence of past states. The random projection approach, also known as sparse spectrum approximation in the Gaussian process community \citep{lazaro2010sparse, wilson2014fast, gal2015improving, dao2017gaussian, li2024trigonometric}. \cite{toth2020bayesian} successfully combined GPs with signatures covariances using an inter-domain inducing point approach aiding in scalability, but leaving the resolution of the quadratic bottleneck in time series length to future work.

\textbf{Deep learning approach.} 
From classic LSTMs \citep{schmidhuber1997long} to transformers \citep{vaswani2017attention}, deep learning methods have been widely applied to sequential data. In the context of probabilistic forecasting, diffusion models \citep{sohl2015deep, ho2020denoising, kollovieh2024predict} are considered among the state-of-the-art techniques. Although deep learning models can approximate any continuous function, they often require a large number of parameters, suffer from high variance, and offer limited interpretability. This creates opportunities for alternative approaches, which can serve not only as competitors but also as complementary components within larger models.

\vspace{-0.5em}
\section{Background}
\vspace{-0.5em}
\subsection{Signature Features and Kernels}
\vspace{-0.5em}
The path signature $S(\bx)$ provides a graded description of a path $\bx:[0,T]\to \bbR^d$ by mapping it to a series of tensors $S(\bx)= \pars{1, S_1(\b x), S_2(\b x), \ldots} \in \prod_{m \geq 0} (\bbR^d)^{\otimes m}$ of increasing degrees, have a long history in many areas of mathematics, such as topology, algebra, and stochastic analysis. 
More recently, it has found applications in machine learning, as a feature map for paths; we refer to \citet[Sec~1.4]{kiraly2019kernels} and \citet{lyons2024signature} for details. 
Among its attractive properties are that: it maps from a nonlinear domain (there is not natural addition of paths of different length) to a linear space $\prod (\bbR^d)^{\otimes m}$; $\bx \mapsto S(\bx)$ is injective up to time-parametrization (thus naturally factoring out time-warping) and it can be made injective by adding time as a path coordinate (i.e.~setting $\bx^0(t)=t)$; it linearizes path functionals, that is we can approximate non-linear path functionals $f(\bx)$ as linear functionals of signatures $f(\bx) \approx \langle \ell, S(\bx)\rangle$; and if the path is random, the expected path signature characterizes the distribution, $\mu \mapsto \mathbb{E}_{\bx \sim \mu}[S(\bx)]$ is injective, see \citet{Chevyrev2018SignatureMT}.
All this makes it a viable transformation for extracting information from time series and paths in machine learning. Truncating the sequence of tensors at a finite level $M \in \bbZ_+$ and using the first $M$ levels as a feature set is common practice. Below we recall its definition and computation.

\vspace{-0.5em}
\textbf{The $m$-th signature level.}
The $m$-th sig.~level is
\begin{align} \label{eq:pathsig_level_m}
    S_m(\bx) = \idotsint\limits_{0 < t_1 < \cdots < t_m < T} \mathrm{d} \bx(t_1) \otimes \cdots \otimes \mathrm{d} \bx(t_m),
\end{align}
resp.~in coordinates $S_m(\bx)$ equals\\
$\left(\,\,\idotsint\limits_{0 < t_1 < \cdots < t_m < T} \mathrm{d} \bx^{i_1}(t_1) \cdots \mathrm{d} \bx^{i_m}(t_m)\right)_{i_1,\ldots,i_m \in \{1,\ldots,d\}}$.
The above integrals make sense as Riemann-Stieljes integrals if we assume $\bx$ is of finite variation, that is for $\bx: [0, T] \to \bbR^d$ a continuous path with finite $1$-variation, $\sup_{\mathcal{D}} \sum_{i=1}^n \norm{\b x(t_{i+1}) - \b x(t_{i})} < \infty$, where $\cD = \curls{t_0 = 0 < t_1 < \cdots < t_n \setgiven n \in \bbN}$ runs over all finite partitions.

\textbf{Discrete-time signature features.}
In practice, we do not have access to continuous paths but are only given a sequence $\bx = (\bx_{0}, \dots, \bx_L) \in \seq(\bbR^d)$  in $\bbR^d$. However, we can simply identify $\bx$ with a continuous time path by its linear interpolation for any choice of time parametrization.
To compute the signature features, denote with $\Delta_m(n)$ to be the collection of all ordered $m$-tuples between $1$ and $n$ such that
\begin{align} \label{eq:delta}
    \Delta_m(n) = \curls{1 \leq i_1 \leq \cdots \leq i_m \leq n}.
\end{align}
 Then, it can be shown using Chen's relation \citep[Thm.~2.9]{lyons2007differential} that its signature can be computed by the recursion for $1 \leq m$, and $1 \leq l \leq L$:
{\small
\begin{align} \label{eq:sig_recursion}
    S_m(\bx_{0:l}) = S_m(\bx_{0:l-1}) + \sum_{p=1}^m S_{m-p}(\bx_{0:l-1}) \otimes \frac{(\delta \bx_l)^{\otimes p}}{p!}
\end{align}
}
with the identification $S_m(\bx_{0:0}) \equiv 0$, and the first-order difference operator is defined as $\delta \bx_k = \bx_k - \bx_{k-1}$. This expression can be evaluated in closed form to get
\begin{align} \label{eq:sigexplicit}
    S_m(\bx_{0:k}) = \sum_{\b i \in \Delta_m(n)} \frac{1}{\b i!} \delta \bx_{i_1} \otimes \cdots \otimes \delta \bx_{i_m},
\end{align}
where $\b i! = k_1! \cdots k_q!$ such that there are $q \in \bbZ_+$ unique indices in the multi-index $\b i$ and $k_1, \cdots, k_q$ are the number of times they are repeated. In other words, $S_m$ summarizes a sequence using the tensor product of contiguous subsequences of length-$m$ normalized by an appropriate factor depending on the number of repetitions for each unique time index. It will be important later on that \eqref{eq:sig_recursion} allows to compute the signature up to all time steps using a paralellizable scan operation. 


\vspace{-0.5em}
\textbf{Signature kernels.} 
As $S_m(\bx)$ is tensor-valued it has $d^m$ coordinates, which makes its computation infeasible for high-dimensional state-spaces or moderately high $m$. 
Kernelization allows to alleviate this by first lifting paths from $\bbR^d$ into a RKHS and subsequently computing inner products of signature features in this RKHS. This alleviates the computational bottleneck and further adds more expressivity by first lifting the paths into a path evolving in a RKHS; see \citet{lee2023signature}. 
Concretely, the signature kernel $\kernel_S: \paths(\bbR^d) \times \paths(\bbR^d) \rightarrow \bbR$ is defined as 
\begin{align} \label{eq:sig_kernel}
    \kernel_S(\bx, \by) = \sum_{m=0}^M \kernel_{S_m}(\bx, \by),
\end{align}
where $M \in \bbZ_+ \cup \{\infty\}$ is the truncation level, and $\b x: [0, S] \to \bbR^d$, $\b y: [0, T] \to \bbR^d$ are continuous paths with finite 1-variation. 
Analogously to \eqref{eq:pathsig_level_m}, we have that
{\small
\begin{align} \label{eq:sig_kernel_m}
    &\kernel_{S_m}(\bx, \by) \\
    =&\idotsint\limits_{\substack{0 < s_1 < \cdots < s_m < S\\0 < t_1 < \cdots < t_m < T}} \inner{\d \kernel_{\bx(s_1)}, \d \kernel_{\by(s_1)}} \cdots \inner{\d \kernel_{\bx(s_m)}, \d \kernel_{\by(t_m)}},
\end{align}}
where $\kernel: \bbR^d \times \bbR^d \to \bbR$ is a base kernel, and for $\bx \in \bbR^d$, $\kernel_\bx \coloneqq \kernel(\bx, \cdot) \in \cH_\kernel$, where $\cH_\kernel$ is the RKHS of $\kernel$. 
Signature kernels and a kernel trick were first introduced in \citet{kiraly2019kernels} for finite $M$ via recursive algorithms, and later for $M=\infty$ via PDE discretization in \citet{salvi2021signature}.
Kernelization allows computing the signature kernel for high- or even infinite-dimensional paths, e.g.~in the setting shown above each path is first lifted into the RKHS $\cH_\kernel$, which is a genuine infinite-dimensional space for non-degenerate kernels $\kernel$. Abstractly, it takes as input a kernel $\kernel$ on the state space $\bbR^d$ and turns it into a kernel $\kernel_S$ for paths in that state space.
The signature kernel is a powerful similarity measure for sequences of possibly different lengths, and it is the state-of-the-art kernel for time series, see e.g.~\citet{toth2023random} for a comprehensive benchmark study and \citet{lee2023signature} for an overview. Although truncating the series \eqref{eq:sig_kernel} may seem like a limitation, it often happens that a careful selection of finite truncation level outperforms the untruncated case, since it can mitigate overfitting.

\textbf{Discrete-time signature kernels.}
Consider two sequences $\bx = (\bx_0, \dots, \bx_K), \by = (\by_0, \dots, \by_L) \in \seq(\bbR^d)$. Similarly to above, we identify each sequence with a path given by its linear interpolation for some choice of time parametrization. Then, substituting into the expression \eqref{eq:sig_kernel_m}, we get the analogous expression to \eqref{eq:sigexplicit}:
\begin{align} \label{eq:discr_sig_kernel}
    &\kernel_{S_m}(\bx, \by) \\
    =& \sum_{\substack{\bi \in \Delta(K)\\ \bj \in \Delta(L)}} \frac{1}{\bi! \bj!} \delta_{i_1, j_1} \kernel(\bx_{i_1}, \by_{j_1}) \cdots \delta_{i_m, j_m} \kernel(\bx_{i_m}, \by_{j_m}),
\end{align}
where $\bi!$ is as previously, and $\delta_{i, j}$ is a second-order differencing operator, such that $\delta_{i, j} \kernel(\bx_i, \by_j) = \kernel(\bx_{i+1}, \by_{j+1}) - \kernel(\bx_{i+1}, \by_j) - \kernel(\b x_i, \b y_{j+1}) + \kernel(\b x_i, \b y_j).$ As being the dual equivalent of \eqref{eq:sigexplicit}, \eqref{eq:discr_sig_kernel} compares two sequences using their contiguous length-$m$ subsequences. A similar recursion to \eqref{eq:sig_recursion} exists to compute \eqref{eq:discr_sig_kernel} that we do not make explicit here, and instead refer to \citet[App.~B]{kiraly2019kernels}, see Algorithm 6. 

\vspace{-0.5em}
\subsection{Random Fourier Features.} 
Kernelization circumvents the computational burden of a high- or infinite-dimensional feature space, but the cost is the associated quadratic complexity in the number of samples which is due to the evaluation of the Gram matrix. 
Random Fourier Features \citep{rahimi2007random} address this by constructing for a given kernel $\kernel$ on $\bbR^d$ a random feature map for which the inner product is a good random approximation to the original kernel. 
Many variations proposed, see the surveys in \citet{liu2021random, chamakh2020orlicz}.

Existing variations are based on Bochner's theorem, which allows for a spectral representation of stationary kernels. Let $\kernel: \bbR^d \times \bbR^d \to \bbR$ be a continuous, bounded, stationary kernel. Then, Bochner's theorem states that there exists a non-negative finite measure $\Lambda$ over $\bbR^d$, such that $\kernel$ is its Fourier transform. In other words, for $\bx, \by, \in \bbR^d$ it holds that $\kernel$ can be be represented as
\begin{align}
    \kernel(\bx, \by) 
    &= \int_{\bbR^d} \exp(i \bs\omega^\top(\bx-\by)) \d \Lambda(\bs\omega)\\ 
    &= \int_{\bbR^d} \cos(\bs\omega^\top (\bx - \by) \d\Lambda(\bs\omega), \label{eq:bochner_cos}
\end{align}
since the kernel is real-valued. Without loss of generality, we may assume that $\Lambda(\bbR^d) = 1$ so $\Lambda$ is a probability measure, since it amounts to rescaling the kernel. Now, we may draw Monte Carlo samples to approximate \eqref{eq:bochner_cos}

\vspace{-10pt}
\begin{align} \label{eq:rff1}
    \kernel(\bx, \by) \approx \frac{1}{D} \sum_{i=1}^{D} \cos(\bs\omega_i^\top(\bx - \by)), 
\end{align}
\vspace{-10pt}

where $D \in \bbZ_+$ and $\bs\omega_1, \dots, \bs\omega_{D} \sim \Lambda$. There are two ways to proceed from here: one is to use the cosine identity $\cos(x - y) = \cos(x)\cos(y) + \sin(x)\sin(y)$, or to further approximate the sum in \eqref{eq:rff1} using the identity
{\small
\begin{align} \label{eq:cos_prob_id}
    \cos(x - y) = \bbE_{b \sim \cU(0, 2\pi)}\bracks{\sqrt{2} \cos(x + b) \sqrt{2} \cos(y + b)}
\end{align}}
see \citet[App.~A]{gal2015improving}. We approximate the expectation \eqref{eq:rff1} by drawing $1$ sample for each term:
\begin{align}
    \kernel(\bx, \by) \approx \frac{2}{D} \sum_{i=1}^{D} \cos(\bs\omega_i^\top \bx + b_i) \cos(\bs\omega_i^\top \by + b_i),
\end{align}
where $b_1, \dots, b_{D} \sim \cU(0, 2\pi)$. This represents each sample using a $1$-dimensional feature, which will be important for us later on in Section \ref{sec:rfsf} when constructing our Random Fourier Signature Feature map. Now, let $\Omega = (\bs\omega_1, \dots, \bs\omega_{D}) \in \bbR^{d \times D}$ and $\bb = (b_1, \dots, b_{D}) \in \bbR^{D}$. The RFF feature map can be written concisely as
\begin{align} \label{eq:rff_feature}
	\tilde\varphi(\b x) = \sqrt{\frac{2}{p}} \cos\pars{\Omega^\top \b x + \b b}.
\end{align}
This offers to approximate a stationary kernel using a random, finite-dimensional feature map, and consequently to reformulate downstream algorithms in weight-space, which avoids the usual cubic costs in the number of data points. Probabilistic convergence of the RFF is studied in the series of works: \citet{rahimi2007random}, optimal rates were derived in \citet{sriperumbudur2015optimal}, extended to the kernel derivatives in \citet{szabo2019kernel}, conditions on the spectral measure relaxed in \citet{chamakh2020orlicz}.  

\section{Recurrent Sparse Spectrum Signature Gaussian Processes} \label{sec:rs3gp}

\subsection{Random Fourier Signature Features } \label{sec:rfsf}
\textbf{Revisited construction.} Now, we detail how RFFs can be combined with signatures to extract random features from a time series. To do so, we revisit the construction of Random Fourier Signature Features (RFSF) from \citet{toth2023random}. We provide an adapted version of the diagonally projected RFSF variant (RFSF-DP), with a twist, which allows us to devise even more computationally efficient features. 


We will focus on the discrete-time setting, but the construction applies to continuous-time in an analogous manner. Firstly, we will build a $1$-dimensional random estimator for the signature kernel. Let $\kernel: \bbR^d \times \bbR^d \to \bbR$ be a continuous, bounded, stationary kernel with spectral measure $\Lambda$, $\bs\omega^{(1)}, \dots, \bs\omega^{(m)} \sim \Lambda$ and $b^{(1)}, \dots, b^{(m)} \sim \cU(0, 2\pi)$, and $\phi^{(p)}(\bx) \coloneqq \cos({\bs\omega^{(p)}}^\top \bx + b^{(p)})$ for $\bx \in \bbR^d$. Then, for sequences $\bx = (\bx_0, \dots, \bx_K), \by = (\by_0, \dots, \by_L) \in \seq(\bbR^d)$, it holds for the signature kernel $\kernel_{S_m}: \seq(\bbR^d) \times \seq(\bbR^d) \to \bbR$ as defined in \eqref{eq:discr_sig_kernel} that
{
\begin{align} \label{eq:sig_estimator}
    &\kernel_{S_m}(\bx, \by) \\ 
    =&\bbE\bracks{\sum_{\substack{\bi \in \Delta_m(K)\\ \bj \in \Delta_m(L)}} \frac{1}{\bi!\bj!} \prod_{p=1}^m \sqrt{2} \delta \phi^{(p)}(\bx_{i_1}) \sqrt{2} \delta \phi^{(p)}(\by_{j_1})}.
\end{align}}
The unbiasedness of the expectation is straightforward to check, since due to linearity and independence of $\phi^{(p)}$'s, it reduces to verifying the relations \eqref{eq:cos_prob_id} and \eqref{eq:bochner_cos}, which in turn recovers the formulation of the signature kernel from \eqref{eq:discr_sig_kernel}. Next, we draw $D \in \bbZ_+$ Monte Carlo samples to approximate \eqref{eq:sig_estimator}, hence, let $\bs\omega^{(1)}_1, \dots, \bs\omega^{(m)}_D \sim \Lambda$ and $b^{(1)}_1, \dots, b^{(m)}_D \sim \cU(0, 2\pi)$, and $\phi^{(p)}_k (\bx) = \cos({\bs\omega^{(p)}_k}^\top \bx + b^{(p)}_k)$ for $\bx \in \bbR^d$. Then,
\begin{align} \label{eq:sig_approx}
    &\kernel_{S_m}(\bx, \by) \\ 
    \approx &\frac{1}{D} \sum_{k=1}^D \sum_{\substack{\bi \in \Delta_m(K)\\ \bj \in \Delta_m(L)}} \frac{1}{\bi!\bj!} \prod_{p=1}^m \sqrt{2} \delta \phi^{(p)}_k(\bx_{i_1}) \sqrt{2} \delta \phi^{(p)}_k(\by_{j_1}).
\end{align}
We call this approximation Random Fourier Signature Features (RFSF), and the features corresponding to the kernel on the RHS of \eqref{eq:sig_approx} can be represented in a finite-dimensional feature space. Let us collect the frequencies $\Omega^{(p)} = (\bs\omega^{(p)}_1, \dots, \bs\omega^{(p)}_D) \in \bbR^{d \times D}$ and phases $\bb^{(p)} = (b^{(p)}_1, \dots, b^{(p)}_D)^\top \in \bbR^D$, and define the RFF maps as in \eqref{eq:rff_feature} so we have $\varphi^{(p)}(\bx) = \cos({\Omega^{(p)}}^\top \bx + \bb^{(p)})$. Then, the RFSF map $\Phi_{m}: \seq(\bbR^d) \to \bbR^D$ is defined for $\bx = (\bx_0, \dots, \bx_K) \in \seq(\bbR^d)$ as
\begin{align} \label{eq:rfsf}
    \Phi_{m}(\bx) = \sqrt{\frac{2^m}{D}} \sum_{\bi \in \Delta_m(K)} \frac{1}{\bi!} \bigodot_{p=1}^m \delta \varphi^{(p)}(\bx_{i_p}),
\end{align}
where $\odot$ refers to the Hadamard product. Although \eqref{eq:rfsf} looks difficult to compute, we can establish a recursion across time steps and signature levels. The update rule analogous to \eqref{eq:sig_recursion} can be written as
\begin{align} \label{eq:rfsf_recursion}
    &\Phi_{m}(\bx_{0:l})\\ 
    =&\Phi_{m}(\bx_{0:l-1}) 
    + \sum_{p=1}^m \frac{1}{p!} \Phi_{m-p}(\bx_{0:l-1}) \bigodot_{q = m-p+1}^m \delta \varphi^{(q)}(\bx_l),
\end{align}
which can be computed efficiently using parallelizable scan operation. Additionally, we overload the definition of the differencing operator $\delta$ to mean fractional differencing for some learnable differencing order $\alpha \in (0, 1)$. Thus, in \eqref{eq:rfsf}, we learn a separate differencing order parameter for each feature channel, see Appendix \ref{app:fracdiff}.

In practive, we use the first $M \in \bbZ_+$ levels in conjunction, and normalize each RFSF level to unit norm, so that the full RFSF map $\Phi: \seq(\bbR^d) \to \bbR^{M D}$ is
\begin{align} \label{eq:rfsf_full}
    \Phi(\bx) = \pars{1, \frac{\Phi_1(\bx)}{\norm{\Phi_1(\bx)}_2}, \dots, \frac{\Phi_M(\bx)}{\norm{\Phi_M(\bx)}_2}}.
\end{align}
We note that $\Phi(\bx)$ can be computed with complexity $O((M+W) L D + M L D d)$, where $W \in \bbZ_+$ is the window size for fractional differencing, see Appendix \ref{app:algs}. This is more efficient than the scalable variants in \citet{toth2023random}, i.e. it avoids the $2^M$ factor as in RFSF-DP, and it also avoids the $D^2$ factor as in RFSF-TRP.

\textbf{Forgetting the past.} The recursive step in RFSF \eqref{eq:rfsf_recursion} allows to compute the feature map up to all time steps of a time series, similarly to a recurrent neural network (RNN), in one forward pass. This suggests to perform sequence-to-sequence regression (including time series forecasting, when the target sequence is future values of the time series) by considering the RFSF map over an expanding window. However, signatures have no built-in forgetting mechanism, and it is well-known that the level-$m$ signature feature has magnitude $O\pars{\nicefrac{\norm{\bx}_\onevar^m}{m!}}$, and the same is true for \eqref{eq:rfsf}, which shows that RFSF only accumulates information without a way of forgetting. This is in contrast to modern RNNs, which have built-in gating mechanisms that allow or disallow the flow of historical information, allowing them to focus on more recent information.

In this section, we propose a novel forgetting mechanism tailored for RFSF, but which can be applied to signature features in general. We tackle this by introducing time step dependent decay factors, which multiply each increment in the formulation \eqref{eq:rfsf}. We assume exponential decay in time, so let $\bs\lambda \in \bbR^D$ be channel-wise decay factors. Then, we define Random Fourier Decayed Signature Features (RFDSF) as
\begin{align} \label{eq:rfdsf}
    &\Phi_{m}(\bx_{0:l})\\ =& \sqrt{\frac{2^m}{D}} \sum_{\bi \in \Delta_m(K)} \frac{1}{\bi!} \bigodot_{p=1}^m \bs\lambda^{\odot(l-i_p)} \odot \delta \varphi^{(p)}(\bx_{i_p}).
\end{align}
A recursion analogous to \eqref{eq:rfsf_recursion} allows to compute \eqref{eq:rfdsf}:
\begin{align} \label{eq:rfdsf_recursion}
    &\Phi_{m}(\bx_{0:l})\\ 
    =&\bs\lambda^{\odot m} \odot \Phi_{m}(\bx_{0:l-1})\\ 
    +& \sum_{p=1}^m \frac{1}{p!} \bs\lambda^{\odot(m-p)} \odot \Phi_{m-p}(\bx_{0:l-1}) \bigodot_{q = m-p+1}^m \delta \varphi^{(q)}(\bx_l),
\end{align}
which can be unrolled over the length of a sequence so
\begin{align} \label{eq:rfdsf_recursion_unrolled}
    &\Phi_{m}(\bx_{0:l})\\ 
    =& \sum_{k=1}^l \bs\lambda^{\odot m(l-k)} \sum_{p=1}^m \frac{1}{p!} \bs\lambda^{\odot(m-p)} \odot \Phi_{m-p}(\bx_{0:l-1})
    \\ &\bigodot_{q = m-p+1}^m \delta \varphi^{(q)}(\bx_k),
\end{align}
which is an exponential decay over time steps of an appropriately chosen sequence depending on signature levels lower than $m$. Hence, \eqref{eq:rfdsf_recursion_unrolled} is still computable by a parallelizable scan operation, allowing for sublinear time computations. This makes the feature map well-suited for long time series, due to the adjustment of context length by the decay factors, and the possibility for parallelization across time. The theoretical complexity is the same as previously, i.e.~$O((M+W)MLD + MLDd)$, see Appendix \ref{app:algs}, but we emphasize that a work-efficient scan algorithm allows to compute the recursion on a GPU log-linearly.

\subsection{Recurrent Sparse Spectrum Signature Gaussian processes}
Next, we construct our (Variational) Recurrent Sparse Spectrum Signature Gaussian Process ((V)RS\textsuperscript{3}GP) model. We define a Gaussian process, which treats the hyperparameters of the random covariance function in a probabilistic fashion \citep{gal2015improving} by incorporating them into the inference procedure. We formulate our variational GP model in the feature space, as opposed to in the function space. 


\textbf{Bayesian formulation.}
We take the setting of supervised learning, and assume our data consists of an input time series $\bx = (\bx_0, \dots, \bx_L) \in \seq(\bbR^d)$ and a univariate output time series $\by = (y_0, \dots, y_L) \in \seq(\bbR)$, although the multivariate case can also be handled by stacking possibly dependent GP priors. 

As the signature is a universal feature map, we expect that a linear layer on top of the computed RFDSF features will work well for approximating functions of time series. We assume a probabilistic model, which models the prediction process as a linear layer on top of RFDSF, corrupted by Gaussian noise, that is,
\begin{align}
    y_l = \b w^\top \Phi(\bx_{0:l}) + \epsilon_l \quad \text{for } l = 0, \dots, L,
\end{align}
where $\b w \in \bbR^{MD}$, such that $M$ is the signature truncation level and $D$ is the RFF dimension. We place an $\iid$ standard Gaussian prior on $\bw$, $p(\b w) = \c N(0, I_D)$. The model noise is $\iid$ Gaussian with learnable variance $\sigma_y^2 > 0$, so that $p(y \given \b w, \bs\Omega, \b B, \bx) = \c N(\b w^\top \Phi(\b x), \sigma_y^2)$, where we explicitly denoted the dependence on the hyperparameters of the RFSF map $\Phi$, frequencies $\bs\Omega = (\Omega^{(1)}, \dots, \Omega^{(M)}) \in \bbR^{M \times d \times D}$, and phases $B = (\b b^{(1)}, \dots, \b b^{(M)}) \in \bbR^{M \times D}$. Next, we specialize to the case of the ARD Gaussian kernel, which has $\Lambda = \cN(\b 0, D^{-1})$, where $D \in \bbR_+^{D \times D}$ is a diagonal matrix of lengthscales. We assign independent lengthscales to each random feature map in \eqref{eq:rfdsf}.
Then, following previous work \citep{gal2015improving, cutajar2017random}, we place the priors on the RFF parameters
\begin{align} \label{eq:rfsf_priors}
&p(\bs\Omega) = \prod_{m=1}^M p(\Omega^{(m)}) = \prod_{m=1}^m \c N^D(\b 0, D_m^{-1}),\\
&p(B) = \prod_{m=1}^M p(\bb^{(m)}) = \prod_{m=1}^M \cU^D(0, 2\pi),
\end{align}
where $D_m = \diag(\ell_{m, 1}, \dots, \ell_{m, d})$ refers to the diagonal lengthscale matrix of the $m^{th}$ RFF map.


As noted in \citep{cutajar2017random}, this model corresponds to a 2-layer Bayesian neural network, where the first layer is given by the RFDSF activations with priors on it as given by \eqref{eq:rfsf_priors}, while the second layer is a linear readout layer with a Gaussian prior on it. Inference in this case is analytically intractable. For a Gaussian likelihood, although $\b w$ can be marginalized out in the log-likelihood, it is not clear how to handle the integrals with respect to $\bs\Omega$  and $B$.  Further, disregarding the latter issue of additional hyperparameters, it only admits full-batch training, and this limits scalability to datasets with large numbers of input-output pairs. To this end, we introduce a variational approximation, which allows for both scalability and flexibility.

\textbf{Variational treatment.}
We variationally approximate the posterior to define an evidence lower bound (ELBO).
We define factorized variational distributions over the parameters $\b w \in \bbR^{MD+1}$, frequencies $\bs\Omega \in \bbR^{M \times d \times D}$, and phases $B \in \bbR^{M \times D}$. The variational over $\b w \in \bbR^D$ is as usual given by a Gaussian $q(\b w) = \c N(\bs\mu_{\b w}, \Sigma_{\b w})$, where $\bs\mu_{\b w} \in \bbR^{MD+1}$ is the variational mean and $\Sigma_{\b w} \in \bbR^{(MD+1) \times (MD+1)}$ is the variational covariance matrix, which is symmetric and positive definite, represented in terms of its Cholesky factor, $L_{\b w} \in \bbR^{(MD+1) \times (MD+1)}$, such that $\Sigma_{\b w} = L_{\b w}L_{\b w}^\top$. The other parameters get the factorized variationals
\begin{align}
	&q(\bs\Omega) = \prod_{m=1}^M q(\Omega^{(m)}) = \prod_{m=1}^M \prod_{i=1}^D \prod_{j=1}^d \c N (\mu_{mij}, \sigma^2_{mij}),\\
    &q(B) = \prod_{m=1}^M q(\b b^{(m)}) =  \prod_{m=1}^M \prod_{i=1}^D \c B_{[0, 2\pi]}(\alpha_{mi}, \beta_{mi}),
\end{align}
where $\alpha_{mi}, \beta_{mi} > 0$, so that the posteriors over the component frequencies of $\omega$ are independent Gaussians, while the posteriors over the phases $b$ are independent beta distributions scaled to $[0, 2\pi]$.

The KL-divergence to the posterior is then minimized:
{
\begin{align}
&\KL{q(\bw, \bs\Omega, B)}{p(\bw, \bs\Omega, B \given \by)}
\\
&= \int q(\b w, \bs\Omega, B) \log \frac{q(\b w, \bs\Omega, B)}{p(\b w, \bs\Omega, B \given \b y)} \d \b w \d \bs\Omega \d B.
\end{align}}

Then, by the usual calculations, we get the Evidence Lower Bound (ELBO), $\c L_{\text{ELBO}} \leq \log p(\b y)$, such that
{\small
\begin{align} \label{eq:elbo}
    \c L_{\text{ELBO}} = &\underbrace{\sum_{i=1}^N \bbE_q\bracks{\log p(y_i \given \b w, \bs\Omega, B)}}_{\text{data-fit term}}\\
    &-\underbrace{\KL{q(\b w)}{p(\b w)} - \KL{q(\bs\Omega)}{p(\bs\Omega)} - \KL{q(B)}{p(B)}}_{\text{KL regularizers}}
\end{align}}

The predictive distribution for a set of (possibly overlapping) sequences $\b X = (\bx_1, \dots, \bx_N) \subset \seq(\bbR^d)$ is then, denoting $\Phi(\b X) = \bracks{\Phi(\bx_1), \ldots, \Phi(\bx_N)}^\top \in \bbR^{N \times (MD+1)}$ and $\b f = \Phi(\b X) \b w \in \bbR^N$,
\begin{align} \label{eq:variational_pred}
    q(\b f) = \c N \left( \Phi(\b X) \bs\mu_{\b w}^\top, \,\, \Phi(\b X) L_{\b w} L_{\b w}^\top \Phi(\b X)^\top \right).
\end{align}

In order to evaluate the data-fit term in \eqref{eq:elbo}, and make inference about unseen points in \eqref{eq:variational_pred}, we require a way to handle the randomness in $\bs\Omega$ and $B$. Two solutions proposed in \citet{cutajar2017random} are to perform Monte Carlo sampling or to sample them once at the start of training and keep them fixed while learning their distributional hyperparameters using the reparametrization trick \citep{kingma2013auto}. Our observations align with theirs, and we found that resampling leads to high variance and non-convergence. The reason for this is likely that the factorized variational distributions do not model correlations between $\bw$ and $\bs\Omega, B$. Hence, we use a fixed random outcome throughout with the reparametrization trick, see Appendix \ref{app:random} for details.

\textbf{Training objective.}
Although the ELBO \eqref{eq:elbo} lower bounds the log-likelihood, it suffers from well-known pathologies. One of these is that it greatly underestimates the latent function uncertainty, and mainly relies on the observation noise for uncertainty calibration, which leads to overestimation. This is clearly unideal for time series forecasting, where heteroscedasticity is often present, and since useful information carried by the kernel about the geometry of the data space is lost. There are several solutions proposed by \citet{jankowiak2020parametric}, and we adopt the approach called Parametric Predictive GP Regression (PPGPR), which restores symmetry between the training objective and the test time predictive distribution by treating both latent function uncertainty and observation noise on equal footing. The key idea is to modify the data-fit term in \eqref{eq:elbo} taking the $\log$ inside the $q$-expectation so we have
{\small
\begin{align}
    \c L_{\text{PPGPR}} =& \sum_{i=1}^N \log\bbE_q\bracks{p(y_i \given \bw, \bs\Omega, B}
    - \KL{q(\b w)}{p(\b w)}
    \\ &- \KL{q(\bs\Omega)}{p(\bs\Omega)} - \KL{q(B)}{p(B)}
\end{align}}
which although is not a lower-bound anymore to the log-likelihood, leads to better calibrated uncertainties. However, as noted by \citet{jankowiak2020parametric}, this objective can lead to underfitting in data regions, where a good fit is harder to achieve, and instead relying on uncertainty overestimation. We account for this by modifying the objective function by introducing a penalty term for the latent function variance. Let $q(f_i) = \cN(\mu_i, \sigma_i^2)$. Then, we modify the objective as
\begin{align}
    \c L =& \sum_{i=1}^N \log\bbE_q\bracks{p(y_i \given \bw, \bs\Omega, B}
    - \KL{q(\b w)}{p(\b w)}
    \\ &- \KL{q(\bs\Omega)}{p(\bs\Omega)} - \KL{q(B)}{p(B)} - \alpha \sum_{i=1}^N \sigma_i^2,
\end{align}
where $\alpha > 0$ is a regularization hyperparameter. This promotes the optimizer to escape local optima, and rely on a better fit of the predictive mean, rather than on uncertainty overestimation to fit the data. We give further details in Appendix \ref{app:obj}.

\section{Experiments}
\textbf{Implementation.} We implemented our models and GP baselines using PyTorch \citep{paszke2019pytorch} and GPyTorch \citep{gardner2018gpytorch}. 
The computing cluster used has 4 NVIDIA RTX 3080 TI GPUs.

\subsection{Synthetic Dataset}

\textbf{Dataset.}
We handcrafted a dataset to test the ability of our (V)RS\textsuperscript{3}GP model of adapting its context length to the data on a task which requires reasoning on multiple time horizons. The dataset consists of a multi-sinusoidal wave with multiple frequencies, which includes both large and low frequencies, requiring reasoning over both short and long time periods. 

\textbf{Methods.} We compare three models: Variational Recurrent Sparse Spectrum Gaussian Process (VRS\textsuperscript{3}GP), our model constructed in the previous section; RS\textsuperscript{3}GP, which ablates the previous model as it does not learn a variational over the random covariance parameters and sets it equal to the prior;  Sparse Variational Gaussian Process (SVGP) using the RBF kernel and a fixed, but hand-tuned context length. (V)RS\textsuperscript{3}GP uses $D = 200$ and $M = 5$, while SVGP uses $100$ inducing points.

\begin{figure}[t]
    \centering
    \includegraphics[width=1.05\hsize]{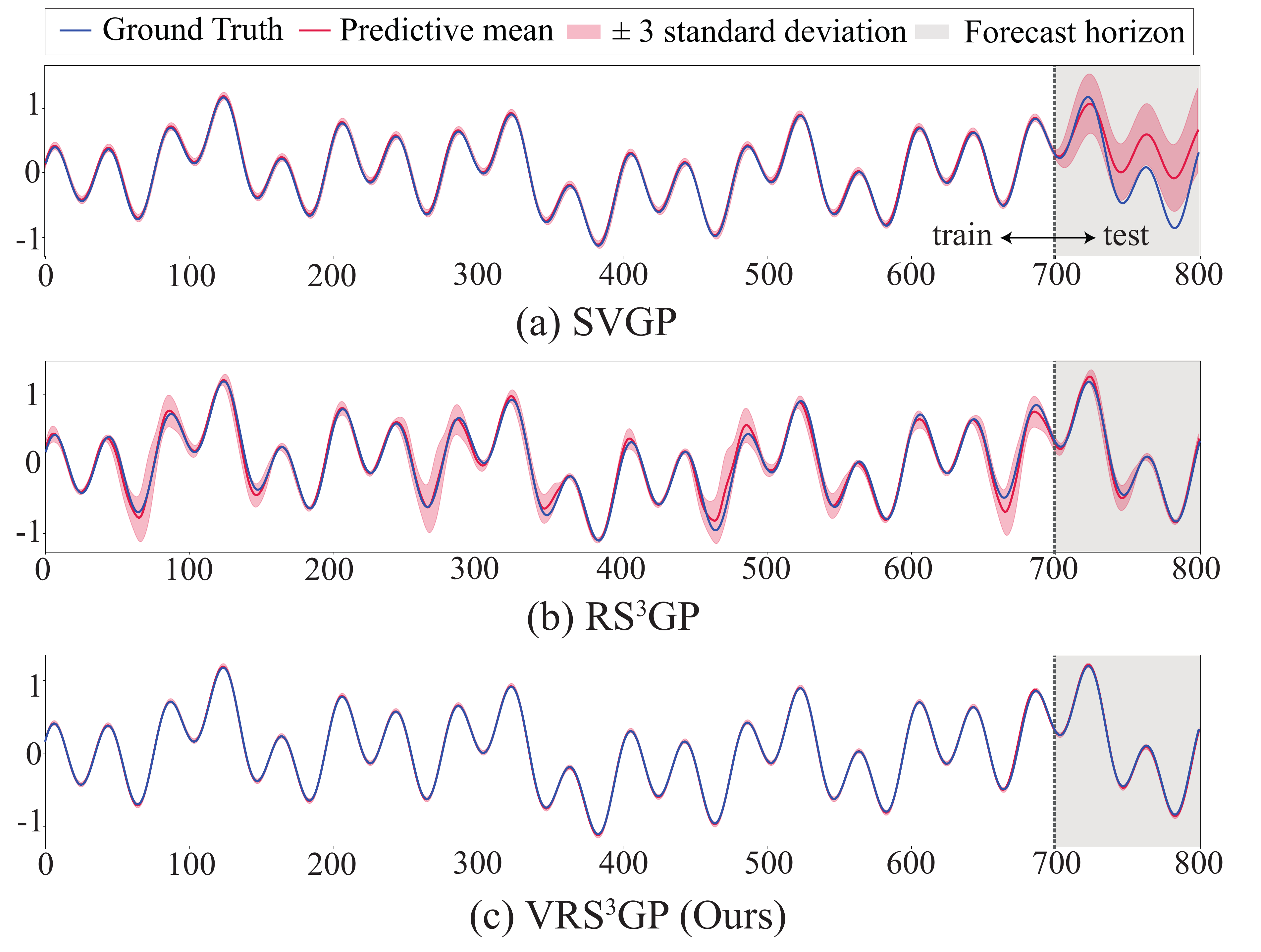}
    \caption{Predictive mean and uncertainty on a toy dataset composed of multi-sinusoidal waves with four distinct components including low and high frequencies, comparing various GP approaches. The true function is depicted in blue, while the predictive mean $\pm 3$ standard deviations are shown in red. The dataset consists of 700 training points, with a context window of 100 for SVGP, and a prediction horizon of 100 for all.}
    \label{fig:demo}
\end{figure}

\textbf{Qualitative analysis.} 
Figure~\ref{fig:demo} qualitatively illustrates the predictive mean and uncertainty of the different approaches. SVGP achieves a perfect fit of the training data, but in the testing regime fails to properly capture the underlying dynamics. RS\textsuperscript{3}GP properly captures both short and long horizon dynamics, but in certain data regions underfits the dataset. VRS\textsuperscript{3}GP on the other hand fits the data perfectly, and also perfectly captures the temporal dynamics, being more flexible than RS\textsuperscript{3}GP due to variational parameter learning. 

\subsection{Real-World Datasets}

\begin{table*}
    \centering
    \caption{Forecasting results on eight benchmark datasets ranked by CRPS. The best and second best models have been shown as \textbf{bold} and \underline{underlined}, respectively.}
    \resizebox{1.0\textwidth}{!}{
    \begin{tabular}{lcccccccc}
        \toprule
         method&  Solar & Electricity & Traffic & Exchange & M4 & UberTLC & KDDCup & Wikipedia\\
         \midrule
         Seasonal Naïve & 0.512 $\pm$ 0.000 & 0.069 $\pm$ 0.000 & 0.221 $\pm$ 0.000 & 0.011 $\pm$ 0.000 & 0.048 $\pm$ 0.000 & 0.299 $\pm$ 0.000 & 0.561 $\pm$ 0.000 & 0.410 $\pm$ 0.000\\
         ARIMA & 0.545 $\pm$ 0.006 & - & - & \textbf{0.008 $\pm$ 0.000} & 0.044 $\pm$ 0.001 & 0.284 $\pm$ 0.001 & 0.547 $\pm$ 0.004 & - \\
         ETS & 0.611 $\pm$ 0.040 & 0.072 $\pm$ 0.004 & 0.433 $\pm$ 0.050 &  \textbf{0.008 $\pm$ 0.000} & 0.042 $\pm$ 0.001 & 0.422 $\pm$ 0.001 & 0.753 $\pm$ 0.008 & 0.715 $\pm$ 0.002 \\
         Linear & 0.569 $\pm$ 0.021 & 0.088 $\pm$ 0.008 & 0.179 $\pm$ 0.003 & 0.011 $\pm$ 0.001 & 0.039 $\pm$  0.001 & 0.360 $\pm$ 0.023 & 0.513 $\pm$ 0.011 & 1.624 $\pm$ 1.114\\
         \midrule
         DeepAR & 0.389 $\pm$ 0.001 & \underline{0.054 $\pm$ 0.000} & \underline{0.099 $\pm$ 0.001} & 0.011 $\pm$ 0.003 & 0.052 $\pm$ 0.006 & \textbf{0.161} $\pm$ \textbf{0.002} & 0.414 $\pm$ 0.027 & 0.231 $\pm$ 0.008 \\
         MQ-CNN & 0.790 $\pm$ 0.063 & 0.067 $\pm$ 0.001 & - & 0.019 $\pm$ 0.006 & 0.046 $\pm$ 0.003 & 0.436 $\pm$ 0.020 & 0.516 $\pm$ 0.012 & 0.220 $\pm$ 0.001 \\
         DeepState & 0.379 $\pm$ 0.002 & 0.075 $\pm$ 0.004 & 0.146 $\pm$ 0.018 & 0.011 $\pm$ 0.001 & 0.041 $\pm$ 0.002 & 0.288 $\pm$ 0.087 & - & 0.318 $\pm$ 0.019 \\
         Transformer & 0.419 $\pm$ 0.008 & 0.076 $\pm$ 0.018 & 0.102 $\pm$ 0.002 & \underline{0.010 $\pm$ 0.000} & 0.040 $\pm$ 0.014 & 0.192 $\pm$ 0.004 & 0.411 $\pm$ 0.021 & \textbf{0.214} $\pm$ \textbf{0.001} \\
         TSDiff & \underline{0.358 $\pm$ 0.020} & \textbf{0.050} $\pm$ \textbf{0.002} & \textbf{0.094} $\pm$ \textbf{0.003} & 0.013 $\pm$ 0.002 &  0.039 $\pm$ 0.006 & \underline{0.172 $\pm$ 0.008} & 0.754 $\pm$ 0.007 & \underline{0.218 $\pm$ 0.010} \\
         \midrule
         SVGP & \textbf{0.341 $\pm$ 0.001} & 0.104 $\pm$ 0.037 & - & 0.011 $\pm$ 0.001 & 0.048 $\pm$ 0.001 & 0.326 $\pm$ 0.043 & 0.323 $\pm$ 0.007 & - \\
         DKLGP & 0.780 $\pm$ 0.269 & 0.207 $\pm$ 0.128 & - & 0.014 $\pm$ 0.004 & 0.047 $\pm$ 0.004 & 0.279 $\pm$ 0.068 & 0.318 $\pm$ 0.010 & - \\
         \midrule
         RS$^3$GP & 0.377 $\pm$ 0.004 & $0.057 \pm 0.001$ & $0.165 \pm 0.001$ & $0.012 \pm 0.001$ & \underline{0.038 $\pm$ 0.003} & $0.354 \pm 0.016$ & \underline{0.297 $\pm$ 0.007} & $0.310 \pm 0.012$ \\
         VRS$^3$GP & $0.366 \pm 0.003$ & 0.056 $\pm$ 0.001 & $0.160 \pm 0.002$ & 0.011 $\pm$ 0.001 & \textbf{0.035} $\pm$ \textbf{0.001} & $0.347 \pm 0.009$ & \textbf{0.291} $\pm$ \textbf{0.015} & $0.295 \pm 0.005$ \\
         \bottomrule
    \end{tabular}
    }
    \label{tab:crps}
\end{table*}
\begin{table}
    \centering
    \caption{Training time in hours}
    \setlength\tabcolsep{3.0pt}
    \resizebox{0.45\textwidth}{!}{
    \begin{tabular}{lccccc}
        \toprule
          & SVGP & DKLGP & TSDiff & RS$^3$GP & VRS$^3$GP\\
         \midrule
         Solar & 1.90 & 2.41 & 2.75 & 0.27 & 0.45\\
         Electricity & 2.23 & 1.91 & 4.09 & 0.62 & 0.98 \\
         Traffic & - & - & 4.44 & 2.31 & 2.70 \\
         Exchange & 0.75 & 0.65  & 2.00 & 0.23 & 0.29 \\
         M4 & 1.68 & 1.94 & 1.96 & 0.86 & 1.02 \\
         UberTLC & 1.89 & 2.19  & 3.19 & 0.53 & 0.71 \\
         KDDCup & 2.03 & 1.55  & 2.10 & 0.65 & 0.93 \\
         Wikipedia & - & - & 6.17 & 3.52 & 6.10 \\
         \bottomrule
    \end{tabular}
    }
    \label{tab:time}
\end{table}
In this section, we present empirical results on several real-world datasets.

\textbf{Datasets.} 
We conducted experiments on eight univariate time series datasets from different domains, available in GluonTS \citep{alexandrov2020gluonts}---Solar \citep{lai2018modeling}, Electricity \citep{asuncion2007uci}, Traffic \citep{asuncion2007uci}, Exchange \citep{lai2018modeling}, M4 \citep{makridakis2020m4}, UberTLC \citep{gasthaus2019probabilistic}, KDDCup \citep{godahewa2021monash}, and Wikipedia \citep{gasthaus2019probabilistic}. We use GluonTS \citep{alexandrov2020gluonts} to load the datasets, which has pre-specified train-test splits for each. See Appendix \ref{app:data} for further details on the datasets.

\textbf{Metric.} We employed the continuous ranked probability score (CRPS) \citep{gneiting2007strictly} for evaluating the quality of probabilistic forecasts. We approximate the CRPS
by the normalized mean quantile loss, and report means and standard deviations over three independent runs, see Appendix \ref{app:metric} for details.

\textbf{Baselines.}
We extend the list of baselines from \citet{kollovieh2024predict}. As such, we have included comprehensive baselines from three groups: classical statistics, deep learning, and other GP methods. For statistics, we included Seasonal Naïve, ARIMA, ETS, and a Linear (ridge) regression model from the statistical literature \citep{hyndman2018forecasting}. Additionally, we compared against deep learning models that represent various architectural paradigms such as the RNN-based DeepAR \citep{salinas2020deepar}, the CNN-based MQ-CNN \citep{wen2017multi}, the state space model-based DeepState \citep{rangapuram2018deep}, the self-attention-based Transformer \citep{vaswani2017attention}, and diffusion-model-based TSDiff \citep{kollovieh2024predict}. For GP models, we add SVGP \citep{hensman2013gaussian}, uses 500 inducing points with the RBF kernel; and DKLGP \citep{wilson2016deep} which augments SVGP with a deep kernel using a $2$-layer NN with $64$ units. See Appendix \ref{app:method} for further details on the methods.

\textbf{Results.} 
Table~\ref{tab:crps} shows the results of our models VRS\textsuperscript{3}GP and RS\textsuperscript{3}GP compared to baselines. We omit results for SVGP and DKLGP on Traffic and Wikipedia since inference time takes longer than 12 hours. Overall, our models achieve top scores on 2 datasets, outperforming the state-of-the-art diffusion baseline TSDiff, and provide comparable performance to deep learning baselines on the remaining 4/6 datasets. Importantly, they outperform linear baselines. They also outperform the GP baselines on 4/6 datasets. One of the datasets our model achieves the top score on, KDDCup, is the one with longest time series length, $L \sim 10000$, which suggests that this is due to its ability to adapt its context memory to long ranges. Moreover, Table~\ref{tab:time} illustrates that our models VRS\textsuperscript{3}GP and RS\textsuperscript{3}GP are significantly quicker to train TSDiff, and both SVGP and DKLGP. This speed up is especially pronounced for datasets with long time series such as Solar, Electricity, Exchange, UberTLC and KDDCup, since our model is able to process a long time series in sublinear time due to parallelizability as detailed in Section \ref{sec:rfsf}. We further investigate the scalability in Appendix \ref{app:scale}.

\section{Conclusion and Limitations}
In this work, we introduced the Random Fourier Decayed Signature Features for time series forecasting, incorporating a novel forgetting mechanism into signature features, which addresses the need to adaptively prioritize local, recent information in long time series. Our proposed model, the Recurrent Sparse Spectrum Signature Gaussian Process (RS\textsuperscript{3}GP), leverages variational inference and recurrent structure to efficiently transform time-series data into a joint predictive distribution. We demonstrated that our approach outperforms traditional GP models and achieves comparable performance to state-of-the-art deep learning methods, while offering significantly faster training times. Limitations include reliance on the Gaussian likelihood, which may be unsuitable for non-Gaussian or heavy-tailed distributions, where asymmetric noise models are warranted, or in cases where explicit target constraints can be used to refine predictions. Additionally, while we introduced a decay mechanism for handling the trade-off between global and local information, more sophisticated forgetting mechanisms might be warranted for capturing complex dependencies in non-stationary time series. Future work could explore these contexts, and extend to the multivariate forecasting regime.

\bibliographystyle{aistats2024_conference}
\bibliography{iclr2025_conference}

\begin{thebibliography}{65}
\providecommand{\natexlab}[1]{#1}
\providecommand{\url}[1]{\texttt{#1}}
\expandafter\ifx\csname urlstyle\endcsname\relax
  \providecommand{\doi}[1]{doi: #1}\else
  \providecommand{\doi}{doi: \begingroup \urlstyle{rm}\Url}\fi

\bibitem[Adachi et~al.(2023)Adachi, Kuhn, Horstmann, Latz, Osborne, and Howey]{adachi2023bayesian}
Masaki Adachi, Yannick Kuhn, Birger Horstmann, Arnulf Latz, Michael~A Osborne, and David~A Howey.
\newblock Bayesian model selection of lithium-ion battery models via {B}ayesian quadrature.
\newblock \emph{IFAC-PapersOnLine}, 56\penalty0 (2):\penalty0 10521--10526, 2023.

\bibitem[Al-Shedivat et~al.(2017)Al-Shedivat, Wilson, Saatchi, Hu, and Xing]{al2017learning}
Maruan Al-Shedivat, Andrew~G Wilson, Yunus Saatchi, Zhiting Hu, and Eric~P Xing.
\newblock Learning scalable deep kernels with recurrent structure.
\newblock \emph{Journal of Machine Learning Research (JMLR)}, 18\penalty0 (82):\penalty0 1--37, 2017.

\bibitem[Alexandrov et~al.(2020)Alexandrov, Benidis, Bohlke-Schneider, Flunkert, Gasthaus, Januschowski, Maddix, Rangapuram, Salinas, Schulz, et~al.]{alexandrov2020gluonts}
Alexander Alexandrov, Konstantinos Benidis, Michael Bohlke-Schneider, Valentin Flunkert, Jan Gasthaus, Tim Januschowski, Danielle~C Maddix, Syama Rangapuram, David Salinas, Jasper Schulz, et~al.
\newblock Gluon{TS}: Probabilistic and neural time series modeling in {P}ython.
\newblock \emph{Journal of Machine Learning Research (JMLR)}, 21\penalty0 (116):\penalty0 1--6, 2020.

\bibitem[Asuncion et~al.(2007)Asuncion, Newman, et~al.]{asuncion2007uci}
Arthur Asuncion, David Newman, et~al.
\newblock {UCI} machine learning repository, 2007.

\bibitem[Bui et~al.(2018)Bui, Pham, Vo, Tran, Nguyen, and Le]{bui2018time}
C~Bui, N~Pham, A~Vo, A~Tran, A~Nguyen, and T~Le.
\newblock Time series forecasting for healthcare diagnosis and prognostics with the focus on cardiovascular diseases.
\newblock In \emph{International Conference on the Development of Biomedical Engineering in Vietnam (BME)}, pp.\  809--818. Springer, 2018.

\bibitem[Chamakh et~al.(2020)Chamakh, Gobet, and Szabó]{chamakh2020orlicz}
Linda Chamakh, Emmanuel Gobet, and Zoltán Szabó.
\newblock Orlicz random {F}ourier features.
\newblock \emph{Journal of Machine Learning Research (JMLR)}, 21\penalty0 (145):\penalty0 1--37, 2020.

\bibitem[Chevyrev \& Oberhauser(2018)Chevyrev and Oberhauser]{Chevyrev2018SignatureMT}
Ilya Chevyrev and Harald Oberhauser.
\newblock Signature moments to characterize laws of stochastic processes.
\newblock \emph{Journal of Machine Learning Research (JMLR)}, 23:\penalty0 176:1--176:42, 2018.
\newblock URL \url{https://api.semanticscholar.org/CorpusID:88523073}.

\bibitem[Cutajar et~al.(2017)Cutajar, Bonilla, Michiardi, and Filippone]{cutajar2017random}
Kurt Cutajar, Edwin~V. Bonilla, Pietro Michiardi, and Maurizio Filippone.
\newblock Random feature expansions for deep {G}aussian processes.
\newblock In \emph{International Conference on Machine Learning (ICML)}, pp.\  884--893, 2017.

\bibitem[Cuturi(2011)]{cuturi2011fast}
Marco Cuturi.
\newblock Fast global alignment kernels.
\newblock In \emph{International Conference on Machine Learning (ICML)}, pp.\  929--936, 2011.

\bibitem[Cuturi \& Doucet(2011)Cuturi and Doucet]{cuturi2011autoregressive}
Marco Cuturi and Arnaud Doucet.
\newblock Autoregressive kernels for time series.
\newblock \emph{arXiv preprint arXiv:1101.0673}, 2011.

\bibitem[Dao et~al.(2017)Dao, De~Sa, and R{\'e}]{dao2017gaussian}
Tri Dao, Christopher~M De~Sa, and Christopher R{\'e}.
\newblock Gaussian quadrature for kernel features.
\newblock \emph{Advances in Neural Information Processing Systems (NeurIPS)}, 30, 2017.

\bibitem[Doerr et~al.(2018)Doerr, Daniel, Schiegg, Duy, Schaal, Toussaint, and Sebastian]{doerr2018probabilistic}
Andreas Doerr, Christian Daniel, Martin Schiegg, Nguyen-Tuong Duy, Stefan Schaal, Marc Toussaint, and Trimpe Sebastian.
\newblock Probabilistic recurrent state-space models.
\newblock In \emph{International Conference on Machine Learning (ICML)}, pp.\  1280--1289. PMLR, 2018.

\bibitem[Eleftheriadis et~al.(2017)Eleftheriadis, Nicholson, Deisenroth, and Hensman]{eleftheriadis2017identification}
Stefanos Eleftheriadis, Tom Nicholson, Marc Deisenroth, and James Hensman.
\newblock Identification of {G}aussian process state space models.
\newblock \emph{Advances in Neural Information Processing Systems (NeurIPS)}, 30, 2017.

\bibitem[Frigola et~al.(2013)Frigola, Lindsten, Sch{\"o}n, and Rasmussen]{frigola2013bayesian}
Roger Frigola, Fredrik Lindsten, Thomas~B Sch{\"o}n, and Carl~Edward Rasmussen.
\newblock Bayesian inference and learning in {G}aussian process state-space models with particle {MCMC}.
\newblock \emph{Advances in Neural Information Processing Systems (NeurIPS)}, 26, 2013.

\bibitem[Frigola et~al.(2014)Frigola, Chen, and Rasmussen]{frigola2014variational}
Roger Frigola, Yutian Chen, and Carl~Edward Rasmussen.
\newblock Variational {G}aussian process state-space models.
\newblock \emph{Advances in Neural Information Processing Systems (NeurIPS)}, 27, 2014.

\bibitem[Gal \& Turner(2015)Gal and Turner]{gal2015improving}
Yarin Gal and Richard Turner.
\newblock Improving the {G}aussian process sparse spectrum approximation by representing uncertainty in frequency inputs.
\newblock In \emph{International Conference on Machine Learning (ICML)}, pp.\  655--664, 2015.

\bibitem[Gardner et~al.(2018)Gardner, Pleiss, Weinberger, Bindel, and Wilson]{gardner2018gpytorch}
Jacob Gardner, Geoff Pleiss, Kilian~Q Weinberger, David Bindel, and Andrew~G Wilson.
\newblock G{P}ytorch: Blackbox matrix-matrix {G}aussian process inference with {GPU} acceleration.
\newblock \emph{Advances in Neural Information Processing Systems (NeurIPS)}, 31, 2018.

\bibitem[Gasthaus et~al.(2019)Gasthaus, Benidis, Wang, Rangapuram, Salinas, Flunkert, and Januschowski]{gasthaus2019probabilistic}
Jan Gasthaus, Konstantinos Benidis, Yuyang Wang, Syama~Sundar Rangapuram, David Salinas, Valentin Flunkert, and Tim Januschowski.
\newblock Probabilistic forecasting with spline quantile function rnns.
\newblock In \emph{International Conference on Artificial Intelligence and Statistics (AISTAS)}, pp.\  1901--1910. PMLR, 2019.

\bibitem[Gneiting \& Raftery(2007)Gneiting and Raftery]{gneiting2007strictly}
Tilmann Gneiting and Adrian~E Raftery.
\newblock Strictly proper scoring rules, prediction, and estimation.
\newblock \emph{Journal of the American statistical Association}, 102\penalty0 (477):\penalty0 359--378, 2007.

\bibitem[Godahewa et~al.(2021)Godahewa, Bergmeir, Webb, Hyndman, and Montero-Manso]{godahewa2021monash}
Rakshitha~Wathsadini Godahewa, Christoph Bergmeir, Geoffrey~I. Webb, Rob Hyndman, and Pablo Montero-Manso.
\newblock Monash time series forecasting archive.
\newblock In \emph{Thirty-fifth Conference on Neural Information Processing Systems Datasets and Benchmarks Track (Round 2)}, 2021.
\newblock URL \url{https://openreview.net/forum?id=wEc1mgAjU-}.

\bibitem[Granger \& Joyeux(1980)Granger and Joyeux]{granger1980introduction}
Clive~WJ Granger and Roselyne Joyeux.
\newblock An introduction to long-memory time series models and fractional differencing.
\newblock \emph{Journal of time series analysis}, 1\penalty0 (1):\penalty0 15--29, 1980.

\bibitem[Hensman et~al.(2013)Hensman, Fusi, and Lawrence]{hensman2013gaussian}
James Hensman, Nicol{\`o} Fusi, and Neil~D Lawrence.
\newblock Gaussian processes for big data.
\newblock In \emph{Uncertainty in Artificial Intelligence (UAI)}, 2013.

\bibitem[Ho et~al.(2020)Ho, Jain, and Abbeel]{ho2020denoising}
Jonathan Ho, Ajay Jain, and Pieter Abbeel.
\newblock Denoising diffusion probabilistic models.
\newblock \emph{Advances in Neural Information Processing Systems (NeurIPS)}, 33:\penalty0 6840--6851, 2020.

\bibitem[Hochreiter \& Schmidhuber(1997)Hochreiter and Schmidhuber]{schmidhuber1997long}
Sepp Hochreiter and J{\"u}rgen Schmidhuber.
\newblock Long short-term memory.
\newblock \emph{Neural Computation}, 9\penalty0 (8):\penalty0 1735--1780, 1997.

\bibitem[Hyndman(2018)]{hyndman2018forecasting}
RJ~Hyndman.
\newblock \emph{Forecasting: principles and practice}.
\newblock OTexts, 2018.

\bibitem[Hyndman \& Khandakar(2008)Hyndman and Khandakar]{hyndman2008automatic}
Rob~J Hyndman and Yeasmin Khandakar.
\newblock Automatic time series forecasting: the forecast package for r.
\newblock \emph{Journal of statistical software}, 27:\penalty0 1--22, 2008.

\bibitem[Ialongo et~al.(2019)Ialongo, Van Der~Wilk, Hensman, and Rasmussen]{ialongo2019overcoming}
Alessandro~Davide Ialongo, Mark Van Der~Wilk, James Hensman, and Carl~Edward Rasmussen.
\newblock Overcoming mean-field approximations in recurrent {G}aussian process models.
\newblock In \emph{International conference on machine learning (ICML)}, pp.\  2931--2940. PMLR, 2019.

\bibitem[Jankowiak et~al.(2020)Jankowiak, Pleiss, and Gardner]{jankowiak2020parametric}
Martin Jankowiak, Geoff Pleiss, and Jacob Gardner.
\newblock Parametric {G}aussian process regressors.
\newblock In \emph{International conference on machine learning (ICML)}, 2020.

\bibitem[Kingma(2014)]{kingma2014adam}
Diederik~P Kingma.
\newblock Adam: A method for stochastic optimization.
\newblock \emph{arXiv preprint arXiv:1412.6980}, 2014.

\bibitem[Kingma \& Welling(2014)Kingma and Welling]{kingma2013auto}
Diederik~P Kingma and Max Welling.
\newblock Auto-encoding variational {B}ayes.
\newblock In \emph{International Conference on Learning Representations (ICLR)}, 2014.

\bibitem[Kir{\'a}ly \& Oberhauser(2019)Kir{\'a}ly and Oberhauser]{kiraly2019kernels}
Franz~J Kir{\'a}ly and Harald Oberhauser.
\newblock Kernels for sequentially ordered data.
\newblock \emph{Journal of Machine Learning Research (JMLR)}, 20\penalty0 (31):\penalty0 1--45, 2019.

\bibitem[Kollovieh et~al.(2024)Kollovieh, Ansari, Bohlke-Schneider, Zschiegner, Wang, and Wang]{kollovieh2024predict}
Marcel Kollovieh, Abdul~Fatir Ansari, Michael Bohlke-Schneider, Jasper Zschiegner, Hao Wang, and Yuyang~Bernie Wang.
\newblock Predict, refine, synthesize: Self-guiding diffusion models for probabilistic time series forecasting.
\newblock \emph{Advances in Neural Information Processing Systems (NeurIPS)}, 36, 2024.

\bibitem[Kong et~al.(2021)Kong, Ping, Huang, Zhao, and Catanzaro]{kong2021diffwave}
Zhifeng Kong, Wei Ping, Jiaji Huang, Kexin Zhao, and Bryan Catanzaro.
\newblock Diffwave: A versatile diffusion model for audio synthesis.
\newblock In \emph{International Conference on Learning Representations}, 2021.
\newblock URL \url{https://openreview.net/forum?id=a-xFK8Ymz5J}.

\bibitem[Lai et~al.(2018)Lai, Chang, Yang, and Liu]{lai2018modeling}
Guokun Lai, Wei-Cheng Chang, Yiming Yang, and Hanxiao Liu.
\newblock Modeling long-and short-term temporal patterns with deep neural networks.
\newblock In \emph{The 41st international ACM SIGIR conference on research \& development in information retrieval}, pp.\  95--104, 2018.

\bibitem[L{\'a}zaro-Gredilla et~al.(2010)L{\'a}zaro-Gredilla, Quinonero-Candela, Rasmussen, and Figueiras-Vidal]{lazaro2010sparse}
Miguel L{\'a}zaro-Gredilla, Joaquin Quinonero-Candela, Carl~Edward Rasmussen, and An{\'\i}bal~R Figueiras-Vidal.
\newblock Sparse spectrum {G}aussian process regression.
\newblock \emph{Journal of Machine Learning Research (JMLR)}, 11:\penalty0 1865--1881, 2010.

\bibitem[Lee \& Oberhauser(2023)Lee and Oberhauser]{lee2023signature}
Darrick Lee and Harald Oberhauser.
\newblock The signature kernel, 2023.

\bibitem[Lemercier et~al.(2021)Lemercier, Salvi, Cass, Bonilla, Damoulas, and Lyons]{lemercier2021siggpde}
Maud Lemercier, Cristopher Salvi, Thomas Cass, Edwin~V Bonilla, Theodoros Damoulas, and Terry~J Lyons.
\newblock Sig{GPDE}: Scaling sparse {G}aussian processes on sequential data.
\newblock In \emph{International conference on machine learning (ICML)}, pp.\  6233--6242. PMLR, 2021.

\bibitem[Li et~al.(2024)Li, Balakirsky, and Mak]{li2024trigonometric}
Kevin Li, Max Balakirsky, and Simon Mak.
\newblock Trigonometric quadrature {F}ourier features for scalable {G}aussian process regression.
\newblock In \emph{International Conference on Artificial Intelligence and Statistics (AISTAS)}, pp.\  3484--3492. PMLR, 2024.

\bibitem[Liu et~al.(2021)Liu, Huang, Chen, and Suykens]{liu2021random}
Fanghui Liu, Xiaolin Huang, Yudong Chen, and Johan~AK Suykens.
\newblock Random features for kernel approximation: A survey on algorithms, theory, and beyond.
\newblock \emph{IEEE Transactions on Pattern Analysis and Machine Intelligence}, 44, 2021.

\bibitem[Lodhi et~al.(2002)Lodhi, Saunders, Shawe-Taylor, Cristianini, and Watkins]{lodhi2002text}
Huma Lodhi, Craig Saunders, John Shawe-Taylor, Nello Cristianini, and Chris Watkins.
\newblock Text classification using string kernels.
\newblock \emph{Journal of Machine Learning Research (JMLR)}, 2\penalty0 (Feb):\penalty0 419--444, 2002.

\bibitem[Lyons \& McLeod(2024)Lyons and McLeod]{lyons2024signature}
Terry Lyons and Andrew~D. McLeod.
\newblock Signature methods in machine learning, 2024.

\bibitem[Lyons \& Oberhauser(2017)Lyons and Oberhauser]{lyons2017sketching}
Terry Lyons and Harald Oberhauser.
\newblock Sketching the order of events.
\newblock \emph{arXiv preprint arXiv:1708.09708}, 2017.

\bibitem[Lyons et~al.(2007)Lyons, Caruana, and L{\'e}vy]{lyons2007differential}
Terry~J Lyons, Michael Caruana, and Thierry L{\'e}vy.
\newblock \emph{Differential equations driven by rough paths}.
\newblock Springer, 2007.

\bibitem[Makridakis et~al.(2020)Makridakis, Spiliotis, and Assimakopoulos]{makridakis2020m4}
Spyros Makridakis, Evangelos Spiliotis, and Vassilios Assimakopoulos.
\newblock The {M4} competition: 100,000 time series and 61 forecasting methods.
\newblock \emph{International Journal of Forecasting}, 36\penalty0 (1):\penalty0 54--74, 2020.

\bibitem[Mattos et~al.(2016)Mattos, Dai, Damianou, Forth, Barreto, and Lawrence]{mattos2016recurrent}
César Lincoln~C. Mattos, Zhenwen Dai, Andreas Damianou, Jeremy Forth, Guilherme~A. Barreto, and Neil~D. Lawrence.
\newblock Recurrent {G}aussian processes.
\newblock In \emph{International Conference on Learning Representations (ICLR)}, volume~3, 2016.

\bibitem[Morrill et~al.(2020)Morrill, Fermanian, Kidger, and Lyons]{morrill2020generalised}
James Morrill, Adeline Fermanian, Patrick Kidger, and Terry Lyons.
\newblock A generalised signature method for multivariate time series feature extraction.
\newblock \emph{arXiv preprint arXiv:2006.00873}, 2020.

\bibitem[Naesseth et~al.(2017)Naesseth, Ruiz, Linderman, and Blei]{naesseth2017reparametrization}
Christian Naesseth, Francisco Ruiz, Scott Linderman, and David Blei.
\newblock Reparameterization gradients through acceptance-rejection sampling algorithms.
\newblock In \emph{International Conference on Artificial Intelligence and Statistics (AISTAS)}, pp.\  489--498, 2017.

\bibitem[Paszke et~al.(2019)Paszke, Gross, Massa, Lerer, Bradbury, Chanan, Killeen, Lin, Gimelshein, Antiga, et~al.]{paszke2019pytorch}
Adam Paszke, Sam Gross, Francisco Massa, Adam Lerer, James Bradbury, Gregory Chanan, Trevor Killeen, Zeming Lin, Natalia Gimelshein, Luca Antiga, et~al.
\newblock Py{T}orch: An imperative style, high-performance deep learning library.
\newblock \emph{Advances in Neural Information Processing Systems (NeurIPS)}, 32, 2019.

\bibitem[Rahimi \& Recht(2007)Rahimi and Recht]{rahimi2007random}
Ali Rahimi and Benjamin Recht.
\newblock Random features for large-scale kernel machines.
\newblock In \emph{Advances in Neural Information Processing Systems (NeurIPS)}, 2007.

\bibitem[Rangapuram et~al.(2018)Rangapuram, Seeger, Gasthaus, Stella, Wang, and Januschowski]{rangapuram2018deep}
Syama~Sundar Rangapuram, Matthias~W Seeger, Jan Gasthaus, Lorenzo Stella, Yuyang Wang, and Tim Januschowski.
\newblock Deep state space models for time series forecasting.
\newblock \emph{Advances in Neural Information Processing Systems (NeurIPS)}, 31, 2018.

\bibitem[Salinas et~al.(2020)Salinas, Flunkert, Gasthaus, and Januschowski]{salinas2020deepar}
David Salinas, Valentin Flunkert, Jan Gasthaus, and Tim Januschowski.
\newblock Deep{AR}: Probabilistic forecasting with autoregressive recurrent networks.
\newblock \emph{International journal of forecasting}, 36\penalty0 (3):\penalty0 1181--1191, 2020.

\bibitem[Salvi et~al.(2021)Salvi, Cass, Foster, Lyons, and Yang]{salvi2021signature}
Cristopher Salvi, Thomas Cass, James Foster, Terry Lyons, and Weixin Yang.
\newblock The signature kernel is the solution of a {G}oursat {PDE}.
\newblock \emph{SIAM Journal on Mathematics of Data Science}, 3\penalty0 (3):\penalty0 873--899, 2021.

\bibitem[Sezer et~al.(2020)Sezer, Gudelek, and Ozbayoglu]{sezer2020financial}
Omer~Berat Sezer, Mehmet~Ugur Gudelek, and Ahmet~Murat Ozbayoglu.
\newblock Financial time series forecasting with deep learning: A systematic literature review: 2005--2019.
\newblock \emph{Applied soft computing}, 90:\penalty0 106181, 2020.

\bibitem[Sohl-Dickstein et~al.(2015)Sohl-Dickstein, Weiss, Maheswaranathan, and Ganguli]{sohl2015deep}
Jascha Sohl-Dickstein, Eric Weiss, Niru Maheswaranathan, and Surya Ganguli.
\newblock Deep unsupervised learning using nonequilibrium thermodynamics.
\newblock In \emph{International conference on machine learning (ICML)}, pp.\  2256--2265. PMLR, 2015.

\bibitem[Sriperumbudur \& Szabo(2015)Sriperumbudur and Szabo]{sriperumbudur2015optimal}
Bharath Sriperumbudur and Zoltan Szabo.
\newblock Optimal rates for random {F}ourier features.
\newblock In \emph{Advances in Neural Information Processing Systems (NeurIPS)}, 2015.

\bibitem[Szabo \& Sriperumbudur(2019)Szabo and Sriperumbudur]{szabo2019kernel}
Zoltan Szabo and Bharath Sriperumbudur.
\newblock On kernel derivative approximation with random {F}ourier features.
\newblock In \emph{International Conference on Artificial Intelligence and Statistics (AISTAS)}, pp.\  827--836, 2019.

\bibitem[Tóth \& Oberhauser(2020)Tóth and Oberhauser]{toth2020bayesian}
Csaba Tóth and Harald Oberhauser.
\newblock Bayesian learning from sequential data using gaussian processes with signature covariances.
\newblock In \emph{International Conference on Machine Learning}, pp.\  9548--9560. PMLR, 2020.

\bibitem[Tóth et~al.(2021)Tóth, Bonnier, and Oberhauser]{toth2021seqtens}
Csaba Tóth, Patric Bonnier, and Harald Oberhauser.
\newblock Seq2{T}ens: An efficient representation of sequences by low-rank tensor projections.
\newblock In \emph{International Conference on Learning Representations (ICLR)}, 2021.
\newblock URL \url{https://openreview.net/forum?id=dx4b7lm8jMM}.

\bibitem[Tóth et~al.(2023)Tóth, Oberhauser, and Szabó]{toth2023random}
Csaba Tóth, Harald Oberhauser, and Zoltan Szabó.
\newblock Random {F}ourier signature features, 2023.

\bibitem[Vaswani et~al.(2017)Vaswani, Shazeer, Parmar, Uszkoreit, Jones, Gomez, Kaiser, and Polosukhin]{vaswani2017attention}
Ashish Vaswani, Noam Shazeer, Niki Parmar, Jakob Uszkoreit, Llion Jones, Aidan~N Gomez, \L~ukasz Kaiser, and Illia Polosukhin.
\newblock Attention is all you need.
\newblock \emph{Advances in Neural Information Processing Systems (NeurIPS)}, 2017.

\bibitem[Wang et~al.(2019)Wang, Lei, Zhang, Zhou, and Peng]{wang2019review}
Huaizhi Wang, Zhenxing Lei, Xian Zhang, Bin Zhou, and Jianchun Peng.
\newblock A review of deep learning for renewable energy forecasting.
\newblock \emph{Energy Conversion and Management}, 198:\penalty0 111799, 2019.

\bibitem[Wen et~al.(2017)Wen, Torkkola, Narayanaswamy, and Madeka]{wen2017multi}
Ruofeng Wen, Kari Torkkola, Balakrishnan Narayanaswamy, and Dhruv Madeka.
\newblock A multi-horizon quantile recurrent forecaster.
\newblock \emph{arXiv preprint arXiv:1711.11053}, 2017.

\bibitem[Williams \& Seeger(2000)Williams and Seeger]{williams2000using}
Christopher Williams and Matthias Seeger.
\newblock Using the {N}ystr{\"o}m method to speed up kernel machines.
\newblock \emph{Advances in Neural Information Processing Systems (NeurIPS)}, 13, 2000.

\bibitem[Wilson et~al.(2014)Wilson, Gilboa, Nehorai, and Cunningham]{wilson2014fast}
Andrew~G Wilson, Elad Gilboa, Arye Nehorai, and John~P Cunningham.
\newblock Fast kernel learning for multidimensional pattern extrapolation.
\newblock \emph{Advances in Neural Information Processing Systems (NeurIPS)}, 27, 2014.

\bibitem[Wilson et~al.(2016)Wilson, Hu, Salakhutdinov, and Xing]{wilson2016deep}
Andrew~G Wilson, Zhiting Hu, Ruslan Salakhutdinov, and Eric~P Xing.
\newblock Deep kernel learning.
\newblock In \emph{Artificial intelligence and statistics (AISTATS)}, pp.\  370--378. PMLR, 2016.

\end{thebibliography}

\newpage
\appendix
\onecolumn

\section{Model details} \label{app:model}

\subsection{Fractional Differencing} \label{app:fracdiff}
Fractional differencing \citep{granger1980introduction} is a technique used in time series analysis to transform non-stationary data into stationary data while preserving long-term dependencies in the series. Unlike traditional differencing methods, which involve subtracting the previous value (integer differencing), fractional differencing allows for a more flexible way of modeling persistence in time series by introducing a fractional order parameter. Hence, fractional differencing generalizes the concept of differencing by allowing the differencing parameter 
$q$ to take on non-integer values.

The general form of fractional differencing is expressed through a binomial expansion:
\begin{align} \label{eq:fracdiff}
    \delta^q X_t = (1 - B)^q X_t = \sum_{k=0}^\infty {q \choose k}(-1)^k X_{t-k},
\end{align}
where $B$ is the backshift operator, $q > 0$ is the fractional differencing order parameter, and $q \choose k$ is the generalized binomial coefficient, i.e.
\begin{align} \label{eq:gen_binom}
    {q \choose k} = \frac{\Gamma(q + 1)}{\Gamma(k+1) \Gamma(q - k + 1)},
\end{align}
where $\Gamma$ is the Gamma function. This formula creates a weighted sum of past values, where the weights decay gradually depending on 
$q$. Note that for integer values of $q$, the formula collapses to standard differencing.

In our case in equation \eqref{eq:rfsf} and \eqref{eq:rfdsf}, we apply fractional differencing to each channel in the lifted time series it is applied to, such that the RFF maps $\varphi^{(p)}$ each have $D$ channels, and each channel is convolved with an individual filter as in \eqref{eq:fracdiff} depending on the channel-wise fractional differencing parameter. In practice, we limit the summation in \eqref{eq:fracdiff} to a finite window $W \in \bbZ_+$, since the sequence decays to zero fast this does not lose much.

\subsection{Algorithms} \label{app:algs}
We adapt the following notation for describing vectorized algorithms from \citet{kiraly2019kernels,toth2021seqtens,toth2023random}. For arrays, $1$-based indexing is used. Elements outside the bounds of an array are treated as zeros as opposed to circular wrapping.
Let $A$ and $B$ be $k$-fold arrays with shape $(n_1 \times \dots \times n_k)$, and let $i_j \in [n_j]$ for $j \in [k]$. We define the following array operations:
\begin{enumerate}[label=(\roman*)]
    \item The slice-wise sum along axis $j$:
    \begin{align}
        A[\dots, :, \Sigma, :, \dots][\dots, i_{j-1}, i_{j+1}, \dots] := \sum_{\kappa=1}^{n_j} A[\dots, i_{j-1}, \kappa, i_{j+1}, \dots].
    \end{align}
    \item  The cumulative sum along axis $j$:
    \begin{align}
        A[\dots, :, \boxplus, :, \dots][\dots, i_{j-1}, i_j,, i_{j+1} \dots] := \sum_{\kappa=1}^{i_j} A[\dots, i_{j-1}, \kappa, i_{j+1}, \dots].
    \end{align}
    \item Let $\bs\lambda \in \bbR^{n_k}$. The channelwise geometric scan along axis $j$:
    \begin{align}
        A[\dots, :, \boxplus^{\bs\lambda}, :, \dots][\dots, i_{j-1}, i_j,, i_{j+1} \dots] := \sum_{\kappa=1}^{i_j} \lambda_{i_k}^{i_j - \kappa} A[\dots, i_{j-1}, \kappa, i_{j+1}, \dots].
    \end{align}
    \item Let $\b q \in \bbR^{n_k}$ and $W \in \bbZ_+$. The channelwise fractional difference along axis $j$:
    \begin{align}
        A[\dots, :, \boxminus^{\b q}_W, :, \dots][\dots, i_{j-1}, i_j,, i_{j+1} \dots] := \sum_{\kappa=0}^{W-1} {q_{i_k} \choose \kappa} A[\dots, i_{j-1}, i_j - \kappa, i_{j+1}, \dots],
    \end{align}
    where ${q_{i_k} \choose \kappa}$ is as in \eqref{eq:gen_binom}.
\item The shift along axis $j$ by $+m$ for $m \in \bbZ+$:
\begin{align}
    A[\dots, :, +m, :, \dots][\dots, i_{j-1}, i_j, i_{j+1}, \dots]
    := A[\dots, i_{j-1}, i_j-m, i_{j+1}, \dots].
\end{align}
\item The Hadamard product of arrays $A$ and $B$:
\begin{align}
 (A \odot B) [i_1, \dots, i_k] := A[i_1, \dots, i_k] B[i_1, \dots, i_k].
\end{align}
\end{enumerate}

\begin{algorithm}[H]
\caption{Computing the RFSF map.}
\label{alg:rfsf}
\begin{algorithmic}[1]
    \STATE {\bfseries Input:} Time series $\bx = (\bx_1, \dots, \bx_L) \in \seq(\bbR^d)$, spectral measure $\Lambda$, truncation $M \in \bbZ_+$, RFF dimension $D \in \bbZ_+$, frac.~diff.~orders $\b q \in \bbR^D$, frac.diff.~window $W \in \bbZ_+$
    \STATE Sample independent RFF frequencies $\Omega^{(1)}, \dots, \Omega^{(M)} \stackrel{\iid}{\sim} \Lambda^{D}$ and phases $\bb^{(1)}, \dots, \bb^{(M)} \stackrel{\iid}{\sim} \cU(0, 2\pi)^D$
    \STATE Initialize an array $U$ with shape $[M, L, D]$
    \STATE Compute RFFs per time step $U[m, l, :] \gets \cos({\Omega^{(m)}}^\top \bx_l + \bb^{(m)})$ for $m \in [M]$ and $l \in [L]$
    \STATE Compute fractional differences $U \gets U[:, \boxminus_W^{\b q}, :]$
    \STATE Accumulate into level-$1$ features $P_1 \gets U[1, \boxplus, :]$
    \STATE Initialize list $R \gets [U[1, :, :]]$
    \FOR{$m=2$ {\bfseries to} $M$}
        \STATE Update with next step $P^\prime \gets P_{m-1}[+1, :] \odot U[m, :, :]$
        \STATE Initialize new list $R^\prime \gets [P^\prime]$
        \FOR{$p=2$ to $m$}
            \STATE Update with current step $Q \gets \frac{1}{p} R[p-1] \odot U[m, :, :]$
            \STATE Append to $R^\prime += [Q]$
        \ENDFOR
        \STATE Aggregate into level-$m$ features $P_m \gets R^\prime[\Sigma][\boxplus, :]$
        \STATE Roll list $R \gets R^\prime$
    \ENDFOR
    \STATE {\bfseries Output:} Arrays of RFSF features per signature level and time step $P_1, \dots, P_M$.
\end{algorithmic}
\end{algorithm}

\begin{algorithm}[H]
\caption{Computing the RFDSF map.}
\label{alg:rfdsf}
\begin{algorithmic}[1]
    \STATE {\bfseries Input:} Time series $\bx = (\bx_1, \dots, \bx_L) \in \seq(\bbR^d)$, spectral measure $\Lambda$, truncation $M \in \bbZ_+$, RFF dimension $D \in \bbZ_+$, frac.~diff.~orders $\b q \in \bbR^D$, frac.diff.~window $W \in \bbZ_+$, channelwise decay factors $\bs\lambda \in \bbR^d$
    \STATE Sample independent RFF frequencies $\Omega^{(1)}, \dots, \Omega^{(M)} \stackrel{\iid}{\sim} \Lambda^{D}$ and phases $\bb^{(1)}, \dots, \bb^{(M)} \stackrel{\iid}{\sim} \cU(0, 2\pi)^D$
    \STATE Initialize an array $U$ with shape $[M, L, D]$
    \STATE Compute RFFs per time step $U[m, l, :] \gets \cos({\Omega^{(m)}}^\top \bx_l + \bb^{(m)})$ for $m \in [M]$ and $l \in [L]$
    \STATE Compute fractional differences $U \gets U[:, \boxminus_W^{\b q}, :]$
    \STATE Accumulate into level-$1$ features $P_1 \gets U[1, \boxplus^{\bs\lambda}, :]$
    \STATE Initialize list $R \gets [U[1, :, :]]$
    \FOR{$m=2$ {\bfseries to} $M$}
        \STATE Update with next step $P^\prime \gets \bs\lambda^{\odot(m-1)} \odot P_{m-1}[+1, :] \odot U[m, :, :]$
        \STATE Initialize new list $R^\prime \gets [P^\prime]$
        \FOR{$p=2$ to $m$}
            \STATE Update with current step $Q \gets \frac{1}{p} R[p-1] \odot U[m, :, :]$
            \STATE Append to $R^\prime += [Q]$
        \ENDFOR
        \STATE Aggregate into level-$m$ features $P_m \gets R^\prime[\Sigma][\boxplus^{\bs\lambda^{\odot m}}, :]$
        \STATE Roll list $R \gets R^\prime$
    \ENDFOR
    \STATE {\bfseries Output:} Arrays of RFDSF features per signature level and time step $P_1, \dots, P_M$.
\end{algorithmic}
\end{algorithm}

\newpage
\subsection{Treatment of Random Parameters} \label{app:random}

\textbf{Reparametrization.}
\begin{figure}
    \centering
    \includegraphics[width=0.5\textwidth]{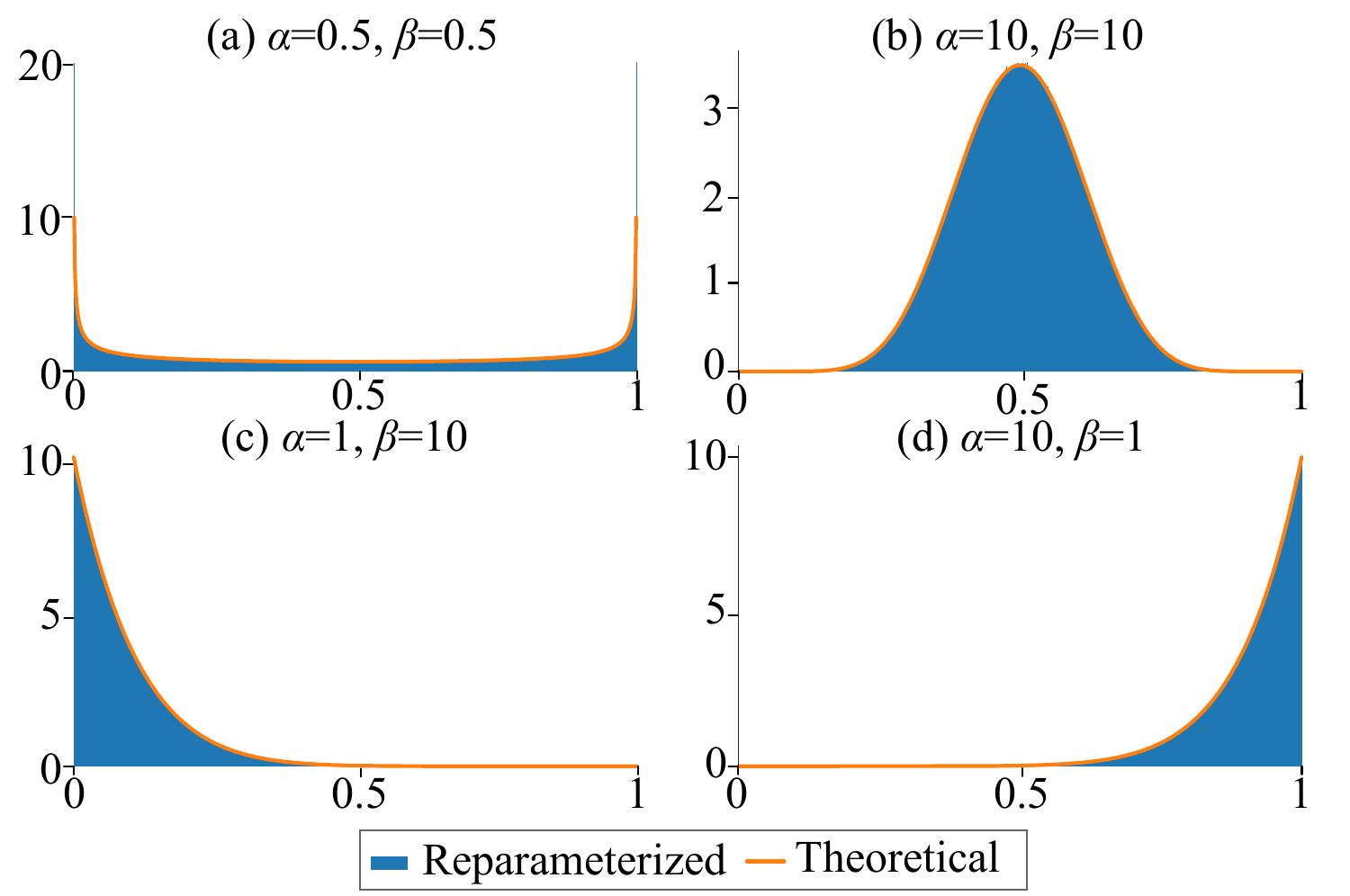}
    \caption{Reparameterizing the beta distribution for various shape parameters given fixed random outcomes.}
    \label{fig:beta_reparam}
\end{figure}
First of all, in order to be able to learn the distributional parameters of the random parameters, we need to reparameterize them \citep{kingma2013auto}. That is, we reparametrize random variable $X$, in terms of a simpler random variable, say $\epsilon$, such that the distributional parameters of $X$ can be captured by a continuous function.  For a Gaussian distribution, this is easy to do, since for $X \sim \c N(\mu, \sigma^2)$, we have the reparametrization $X = \sigma \epsilon + \mu$, where $\epsilon \sim \c N(0, 1)$. For a beta-distributed random variable, this is more tricky, and it is unfeasible to do exact reparameterization. The problem is equivalent to reparameterizing a Gamma random variable, since for $X = Y_1 / (Y_1 + Y_2)$ such that $Y_1 \sim \Gamma(\alpha, 1)$, $Y_2 \sim \Gamma(\beta, 1)$, it holds that $X \sim \c B(\alpha, \beta)$. Previous work \citep{naesseth2017reparametrization} combined accepted samples from the Gamma acceptance-rejection sampler with the shape augmentation trick for Gamma variables, so that the probability of acceptance is close to 1 in order to reparameterize Gamma random variables. We numerically validated this approach, and for fixed randomness, it is able to continuously transport between beta distributions with different shape parameters. The below figure shows a density plot of the reparameterization for shape augmentation parameter $B = 10$ and $n = 10^7$ samples. We see that the theoretical densities are reproduced exactly by the reparameterized distributions.

\textbf{Resample or not to resample.}
The evaluation of the KL divergences can be done analytically, and the question becomes how to sample from the data-fit term, and how to treat the integrals with respect to the random parameters during inference. There are two approaches proposed in \citet{cutajar2017random}. First, is to simply apply vanilla Monte Carlo sampling to the random hyperparameters, that is, resample them at each step optimization step. This leads to high variance and slow convergence of the ELBO, or it may not even converge within optimization budgets. The explanation given for this is that the since the variational factorizes across the ``layers'' of the model, it does not maintain the proper correlations needed to propagate the uncertainty. Hence, instead of resampling the random parameters during training, we retort to sampling them once at the start of training, and then keeping them fixed using the reparametrization trick throughout training and then inference.

\subsection{Fixing Pathologies in the ELBO} \label{app:obj}
Next, we will focus on why the ELBO is unsuitable for capturing latent function uncertainty, and relies solely on the observation noise for uncertainty calibration. Let us recall the ELBO from \eqref{eq:elbo}:
{\begin{align}
    \c L_{\text{ELBO}} = &\underbrace{\sum_{i=1}^N \bbE_q\bracks{\log p(y_i \given \b w, \bs\Omega, B)}}_{\text{data-fit term}} -\underbrace{\KL{q(\b w)}{p(\b w)} - \KL{q(\bs\Omega)}{p(\bs\Omega)} - \KL{q(B)}{p(B)}}_{\text{KL regularizers}}.
\end{align}}
Clearly, the KL regularizers do not play a role in uncertainty calibration to the data, and we will focus on a datafit term $\bbE_q\bracks{\log p(y_i \given \bw, \bs\Omega, B)}$. Let $f_i = \bw^\top \Phi(\bx_i)$, and $q(f_i) = \cN(\mu_i, \sigma_i)$ be given as in \eqref{eq:variational_pred}. Now, note that $p(y_i \given \bw, \bs\Omega, B) = \cN(y_i \given f_i, \sigma^2_y)$. Hence,
\begin{align}
    \bbE_q\bracks{\log p(y_i \given \bw, \bs\Omega, B)} = - \frac{1}{2} \log(2\pi \sigma_y^2) - \frac{1}{2 \sigma_{y}^2} \bbE_q[y_i - f_i]^2. 
\end{align}
The first term clearly plays no role in the data calibration, but serves as a penalty, while the second term matches the predictive distribution to the data. Calculating the expectation, we get that
\begin{align}
    \frac{1}{2\sigma_y^2}\bbE_q\bracks{y_i - f_i}^2 = \frac{1}{2\sigma_y^2}\pars{(y_i - \mu_i)^2 + \sigma_i^2}.
\end{align}
Putting these two together, we get that
\begin{align} \label{eq:elbo_expanded}
    bbE_q\bracks{\log p(y_i \given \bw, \bs\Omega, B)} = - \frac{1}{2} \log(2\pi \sigma_y^2) - \frac{1}{2 \sigma_{y}^2} \pars{(y_i- \mu_i)^2 + \sigma_i^2}.
\end{align}
Hence, the data-fit is solely determined by the closeness of the predictive mean to the underlying observation weighted the inverse of the observation noise level, while the latent function uncertainty is simply penalized without playing a role in the data-fit calibration. This explains why the ELBO underestimates the latent function uncertainty and only relies on the observation noise for uncertainty calibration.

Next, we consider the PPGPR loss. With the notation as above, consider the modification of the ELBO
{\begin{align}
    \c L_{\text{PPGPR}} = &\underbrace{\sum_{i=1}^N \log \bbE_q\bracks{p(y_i \given \b w, \bs\Omega, B)}}_{\text{data-fit term}} -\underbrace{\KL{q(\b w)}{p(\b w)} - \KL{q(\bs\Omega)}{p(\bs\Omega)} - \KL{q(B)}{p(B)}}_{\text{KL regularizers}}.
\end{align}}
Now, the data-fit term is given by the $\log$ outside of the $q$-expectation so that $\bbE_q[p(y_i \given \bw, \bs\Omega, B)] = \bbE_q[\cN(y_i \given f_i, \sigma_y^2)]$, which is a convolution of two Gaussians, hence we get that
\begin{align} \label{eq:ppgpr_expanded}
    \log \bbE_q[p(y_i \given \bw, \bs\Omega, B)] = \log \cN(y_i \given \mu_i, \sigma_f^2 + \sigma_y^2) = -\frac{1}{2}\log(2\pi(\sigma_i^2 + \sigma_y^2)) - \frac{1}{2(\sigma_i^2 + \sigma_y^2)} (y_i - \mu_i)^2. 
\end{align}
Contrasting this with \eqref{eq:elbo_expanded}, we see that \eqref{eq:ppgpr_expanded} treats the latent function variance $\sigma_i^2$ and $\sigma_y^2$ on equal footing, hence restoring symmetry in the objective function. This is why PPGPR often leads to better predictive variances, since both the latent function uncertainty and observation noise are taken into account during uncertainty calibration, which makes use of the expressive power contained in the kernel function regarding the geometry of the data-space.

\section{Experiment Details} \label{app:exp}
\subsection{Datasets} \label{app:data}
\begin{table}
    \centering
    \caption{Dataset description}
    \setlength\tabcolsep{3.0pt}
    \resizebox{1\textwidth}{!}{
    \begin{tabular}{llccccccc}
        \toprule
         Dataset & GluonTS Name & Train Size & Test Size & Domain & Freq. & Median Seq. Length & Context Length & Prediction Length\\
         \midrule
         Solar & \texttt{solar\_nips} & 137 & 959 & $\mathbb{R}^+$ & H & 7009 & 336 & 24\\
         Electricity & \texttt{electricity\_nips} & 370 & 2590 & $\mathbb{R}^+$ & H & 5833 & 336 & 24\\
         Traffic & \texttt{traffic\_nips} & 963 & 6741 & (0,1) & H & 4001 & 336 & 24\\
         Exchange & \texttt{exchange\_rate\_nips} & 8 & 40 & $\mathbb{R}^+$ & D & 6071 & 360 & 30\\
         M4 & \texttt{m4\_hourly} & 414 & 414 & $\mathbb{N}$ & H & 960 & 312 & 48\\
         KDDCup & \texttt{kdd\_cup\_2018\_without\_missing} & 270 & 270 & $\mathbb{N}$ & H & 10850 & 312 & 48\\
         UberTLC & \texttt{uber\_tlc\_hourly} & 262 & 262 & $\mathbb{N}$ & H & 4320 & 336 & 24\\
         Wikipedia & \texttt{wiki2000\_nips} & 2000 & 10000 & $\mathbb{N}$ & D & 792 & 360 & 30\\
         \bottomrule
    \end{tabular}
    }
    \label{tab:dataset}
\end{table}
We conducted our experiments using eight widely recognized univariate datasets from various domains. Preprocessed versions of these datasets can be found in GluonTS \citep{alexandrov2020gluonts}, which includes details on their frequencies (either daily or hourly) and corresponding prediction lengths. An overview of these datasets is presented in Table~\ref{tab:dataset}. For the hourly datasets, we utilized sequence lengths of 360 (equivalent to 15 days), while the daily datasets had sequence lengths of 390 (approximately 13 months). These sequences were created by randomly slicing the original time series at different timesteps. Below, we provide a brief description of each dataset.

\begin{enumerate}
    \item The Solar dataset \citep{lai2018modeling} includes data on photovoltaic power generation from 137 solar power plants in Alabama for the year 2006.
    \item The Electricity dataset \citep{asuncion2007uci} features consumption records from 370 customers.
    \item The Traffic dataset \citep{asuncion2007uci} provides hourly occupancy statistics for freeways in the San Francisco Bay area from 2015 to 2016.
    \item The Exchange dataset \citep{lai2018modeling} contains daily exchange rate information for eight countries, including Australia, the UK, Canada, Switzerland, China, Japan, New Zealand, and Singapore, spanning from 1990 to 2016.
    \item M4 \citep{makridakis2020m4} refers to a subset of hourly data from the M4 forecasting competition.
    \item The KDDCup dataset \citep{godahewa2021monash} comprises air quality indices (AQIs) from Beijing and London, utilized during the KDD Cup 2018.
    \item The UberTLC dataset \citep{gasthaus2019probabilistic} consists of Uber pickup records collected between January and June 2015, sourced from the New York City Taxi and Limousine Commission (TLC). 
    \item the Wikipedia dataset \citep{gasthaus2019probabilistic} includes daily counts of visits for 2,000 specific Wikipedia pages.
\end{enumerate}

\subsection{Assets} \label{app:assets}
We used the following assets for the implementation. The licences of the assets are as follows:
\begin{enumerate}
    \item PyTorch \citep{paszke2019pytorch} only claims the copyright by Facebook, Inc (Adam Paszke).
    \item GPyTorch \citep{gardner2018gpytorch} is under MIT license.
    \item GluonTS \citep{alexandrov2020gluonts} is under Apache-2.0 license.
    \item accelerated-scan\footnote{\href{https://github.com/proger/accelerated-scan}{https://github.com/proger/accelerated-scan}} under MIT license.
\end{enumerate}
For the dataset, the authors allow us to use for academic publication purpose if we cite their original papers. We have cited corresponding papers one by one.

\subsection{Metric} \label{app:metric}
We employed 
the continuous ranked probability score (CRPS; \citet{gneiting2007strictly}), 
\begin{align}
    \mathrm{CRPS}(F^{-1}, y) = \int_0^1 2 \mathcal{L}_\tau \left( F^{-1}(\tau), y \right) d \tau,
\end{align}
where $\mathcal{L}_\tau(q, y) = (\tau - \{y < q\})(y - q)$ is the pinball loss for a specific quantile level $\tau$, $F^{-1}$ is the predicted inverse cumulative distribution function (also known as the quantile function).
Unlike the MSE, which only measures the average squared deviation from the mean, the CRPS captures the entire distribution by integrating the pinball loss across all quantiles, from 0 to 1. 
Although GP can offer an analytical CRPS calculation due to its Gaussianity, we adopted a common empirical approximation using GluonTS \citep{alexandrov2020gluonts}, for the fair comparison with existing literature including non-GP-based approaches. In GluonTS, the CRPS is approximated using discrete quantile levels derived
from samples, which defaults to nine quantile levels,
$\{0.1,0.2,0.3,0.4,0.5,0.6,0.7,0.8,0.9\}$.
\subsection{Methods} \label{app:method}
\textbf{Frequentist methods.}
We detail the frequentist methods we borrowed results from \citet{kollovieh2024predict}.
\begin{itemize}
    \item Seasonal Naive approach \citep{hyndman2018forecasting} is a straightforward forecasting technique that predicts future values by using the most recent seasonal observation. For example, in a time series with hourly data, it utilizes the value from the same hour the previous day.
    \item ARIMA \citep{hyndman2018forecasting} is a widely-used statistical model for time series analysis and forecasting. It integrates autoregressive (AR), differencing (I), and moving average (MA) elements to effectively model and predict trends and variations. The implementation is based on the forecast package \citep{hyndman2008automatic} for R.
    \item ETS \citep{hyndman2018forecasting} is another forecasting technique that leverages exponential smoothing to account for trend, seasonality, and error components within the time series data. The implementation is based on the forecast package \citep{hyndman2008automatic} for R.
    \item Linear \citep{hyndman2018forecasting} refers to a ridge regression model that operates over time, using the previous context of a specified number of timesteps as inputs to forecast the subsequent timesteps. The implementation is based on the forecast package \citep{hyndman2008automatic} for R, applied the default regularization strength of 1. The model was trained on 10,000 randomly selected sequences, maintaining the same context length. 
    \item DeepAR \citep{salinas2020deepar} is an autoregressive model based on recurrent neural networks (RNNs) that incorporates the historical context of the time series through lags and additional relevant features, including temporal elements like hour of the day and day of the week. The model outputs the parameters for the next distribution (such as Student’s-t or Gaussian) in an autoregressive manner and is optimized to maximize the conditional log-likelihood. The implementation is based on GluonTS \citep{alexandrov2020gluonts}, following the suggested hyperparameter settings.
    \item MQ-CNN \citep{wen2017multi} employs a convolutional neural network (CNN) framework to identify patterns and relationships in time series data. This model is optimized using quantile loss, enabling it to produce multi-horizon forecasts directly. The implementation is based on GluonTS \citep{alexandrov2020gluonts}, adhering to the suggested hyperparameters.
    \item DeepState \citep{rangapuram2018deep} integrates recurrent neural networks (RNNs) with linear dynamical systems (LDS). In this model, the RNN processes additional input features, such as time attributes and item identifiers, to generate the diagonal noise covariance matrices for the LDS. Key parameters of the LDS, including the transition and emission matrices, are crafted manually to represent various components of time series data, including level, trend, and seasonality. The training of both the LDS and RNN occurs through maximum likelihood estimation. The implementation is based on GluonTS \citep{alexandrov2020gluonts}, adhering to the suggested hyperparameter settings.
    \item Transformer \citep{vaswani2017attention} serves as a sequence-to-sequence forecasting tool that employs a self-attention mechanism. It accepts time series lagged values and covariates as inputs, producing parameters for future probability distributions. The implementation is based on GluonTS \citep{alexandrov2020gluonts}, following the recommended hyperparameter configurations.
    \item TSDiff is a conditional diffusion model for univariate time series imputation and forecasting. TSDiff is based on the DiffWave architecture \citep{kong2021diffwave}, integrating the observed context together with an observation mask using a Conv1x1 layer followed by an addition after the S4 layer in each residual block. The implementation is based on their original implementation\footnote{\url{github.com/amazon-science/unconditional-time-series-diffusion}} with the recommended hyperparameters and training setup.
\end{itemize}

\newpage
\textbf{GP methods.} Here we describe the GP baselines and our methods.
\begin{itemize}
    \item SVGP and DKLGP both use a Sparse Variational Gaussian Process base model \citep{hensman2013gaussian} using $n_Z = 500$ inducing points and using the RBF kernel. DKLGP augments the kernel with a $2$-layer neural network with $64$ units. Both models are trained in an autoregressive way getting as input the previous context length number of time steps and predict the next step prediction length ahead. We used Adam \citep{kingma2014adam} optimizer with a batch size of $n=128$, a learning rate of $\alpha = 10^{-3}$, and for $200$ epochs, but the number of training steps were capped above $2 \times 10^5$ steps to keep the training time comparable.
    \item RS\textsuperscript{3}GP and VRS\textsuperscript{3}GP both use our RFDSF feature map trained in an autoregressive way receiving the entire observed history until a given point in time as input and predicting prediction length number of steps ahead. RS\textsuperscript{3}GP ablates the variational distributions on the RFF parameters and sets them equal to the prior. For both models, we augment the input univariate time series using $l=9$ of its lags (shifted values in time), so that the time series the model operates on is $d = 10$ dimensional. The hyperparameters used were $D = 200$ and $M=5$. The batch size is set to $1$ and each batch contains a full time series from the training set, which is processed in one pass and transformed into the corresponding predictive distributions over the time steps. For training Adam \citep{kingma2014adam} was used with a learning rate of $\alpha = 10^{-3}$, and for $200$ epochs, but for at least $2 \times 10^4$ number of steps to avoid underfitting on smaller datasets.
\end{itemize}
Additionally, for each GP model, both baselines and our models, we perform an additional uncertainty calibration step per time series during inference. For each time series in the test set, we multiply the predictive standard deviations by a scalar $\beta \in [0.1, 0.2, \dots, 2]$, where $\beta$ is selected to maximize the CRPS on the observed part of the time series, i.e.~the time series length minus the last prediction length steps, where the actual testing happens. This allows to fine-tune predictions in a forward looking bias free way, and often yields improvements in results.

\subsection{Scalability} \label{app:scale}

\begin{figure}[h]
\begin{minipage}{0.49\textwidth}
    \includegraphics[width=\textwidth]{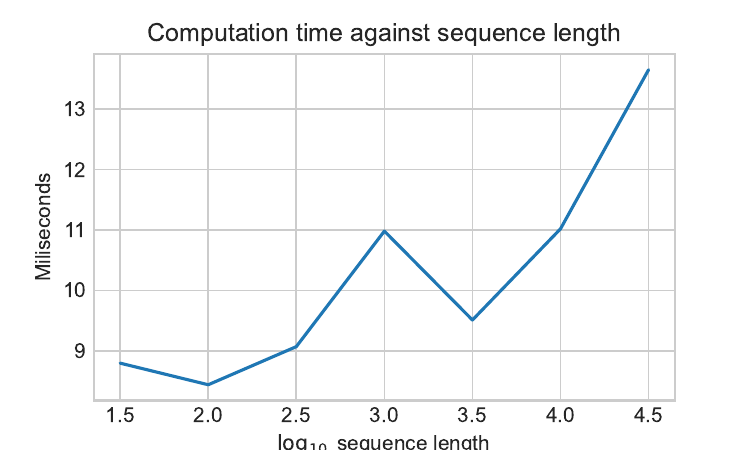}
\end{minipage}
\begin{minipage}{0.49\textwidth}
    \includegraphics[width=\textwidth]{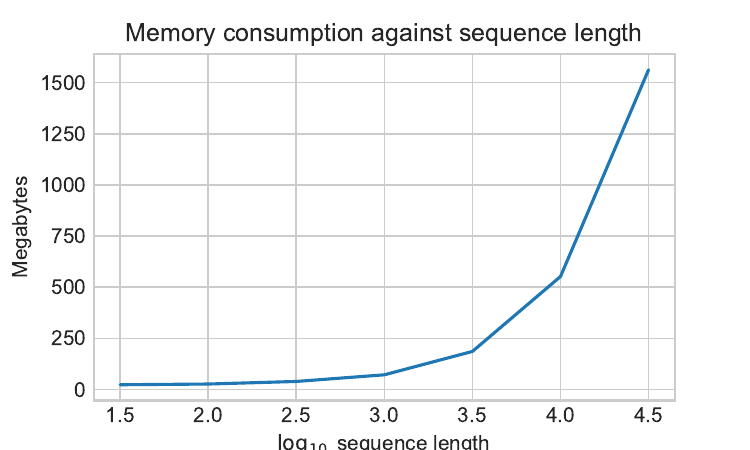}
\end{minipage}
\caption{Computation time (left) and memory consumption (right) of VRS\textsuperscript{3}GP measured against scaling the sequence length on a logarithmic scale. The hyperparameters of the model are $D = 200$ and $M = 5$, and the input time series is univariate augmented with $l = 9$ lags.}
\label{fig:benchmark}
\end{figure}


\end{document}